\documentclass[10pt,journal]{IEEEtran}
\usepackage{epsfig}
\usepackage{float}
\usepackage{graphicx}
\usepackage{amsmath}
\usepackage{amssymb}
\usepackage{amsthm}
\usepackage[linesnumbered,ruled]{algorithm2e}
\usepackage{colortbl,booktabs}
\usepackage{threeparttable}
\usepackage{subeqnarray}
\usepackage{cases}
\usepackage{multirow}
\usepackage{array}
\usepackage[table]{xcolor}
\usepackage{bm}
\usepackage[switch]{lineno}
\usepackage{threeparttable}
\usepackage{fixltx2e}
\usepackage{enumerate}
\usepackage{epstopdf}
\usepackage[hyphenbreaks]{breakurl}
\usepackage[hyphens]{url}
\usepackage{indentfirst}
\usepackage{color}
\newcolumntype{I}{!{\vrule width 2pt}}
\newlength\savedwidth

\newlength\savewidth

\theoremstyle{definition}

\ifCLASSINFOpdf
\else
\fi
\begin{document}
\title{Cross-Domain Label Propagation for Domain Adaptation  with Discriminative Graph  Self-Learning}
%
%
%

\author{Lei Tian*,~
        Yongqiang Tang*,~
		Liangchen Hu
		and Wensheng~Zhang~
\thanks{* indicates equal contributions.}
\thanks{L. Tian, and W. Zhang are with the Research Center of Precision Sensing and Control, Institute of Automation, Chinese Academy of Sciences, Beijing, 100190, China, and the School of Artificial Intelligence, University of Chinese Academy of Sciences, Beijing, 101408, China.
E-mail:\{tianlei2017@ia.ac.cn, zhangwenshengia@hotmail.com.}
\thanks{Y. Tang is with the Research Center of Precision Sensing and Control, Institute of Automation, Chinese Academy of Sciences, Beijing, 100190, China.
E-mail:yongqiang.tang@ia.ac.cn}
\thanks{L. Hu is with the School of Computer Science and Engineering, Nanjing University of Science and Technology, Nanjing, 210094, China.
	E-mail:hlc\_clear@foxmail.com.}
}

%
%

\markboth{}%
{Shell \MakeLowercase{\textit{et al.}}: Bare Demo of IEEEtran.cls for IEEE Journals}
%



\maketitle

\begin{abstract}
Domain adaptation manages to transfer the knowledge of well-labeled source data to unlabeled target data. Many recent efforts focus on improving the prediction accuracy of target pseudo-labels to reduce conditional distribution shift. In this paper, we propose a novel domain adaptation method, which infers target pseudo-labels through cross-domain label propagation, such that the underlying manifold structure of two domain data can  be explored. Unlike existing cross-domain label propagation methods that separate domain-invariant feature learning, affinity matrix constructing and target labels inferring into three independent stages, we propose to integrate them into a unified optimization framework. In such way, these three parts can boost each other  from an iterative optimization perspective and thus more effective knowledge transfer can be achieved. Furthermore, to construct a high-quality affinity matrix, we propose a discriminative graph self-learning strategy, which can  not only
adaptively capture the inherent similarity of the data from two domains but also effectively exploit the  discriminative information contained in well-labeled source data and pseudo-labeled target data.
An efficient iterative optimization algorithm is designed  to solve the objective function of our proposal. Notably, the proposed method can be  extended to semi-supervised domain adaptation in a simple but effective way and the corresponding optimization problem can be solved with the identical   algorithm. Extensive experiments on six standard datasets verify the significant superiority of our proposal in both unsupervised and semi-supervised domain adaptation settings.
\end{abstract}


%
\IEEEpeerreviewmaketitle

\section{Introduction \label{Introduction}}
%
%
%
%

\IEEEPARstart{O}{ne} common assumption of statistical learning theory is that the training data and test data are drawn from an identical feature distribution, which may be violated in many situations. Moreover, in practical applications, collecting labeled training data is often expensive and time-consuming. Thus, there is a strong demand to leverage the knowledge from a source domain with sufficient labels to help design effective model for the unlabeled target domain data, which follows a different  feature distribution. To this end, considerable efforts have been devoted to domain adaptation \cite{Survey}, and impressive progress has been made in various tasks, {\it e.g.}, object recognition \cite{ MCTL,HCA, CMMS}, semantic segmentation \cite{ReID,ReID1},  and sentiment analysis \cite{PRDA,SDAF}.

\begin{figure*}[h]
	\setlength{\abovecaptionskip}{0pt}
	\setlength{\belowcaptionskip}{0pt}
	\renewcommand{\figurename}{Figure}
	\centering
	\includegraphics[width=0.98\textwidth]{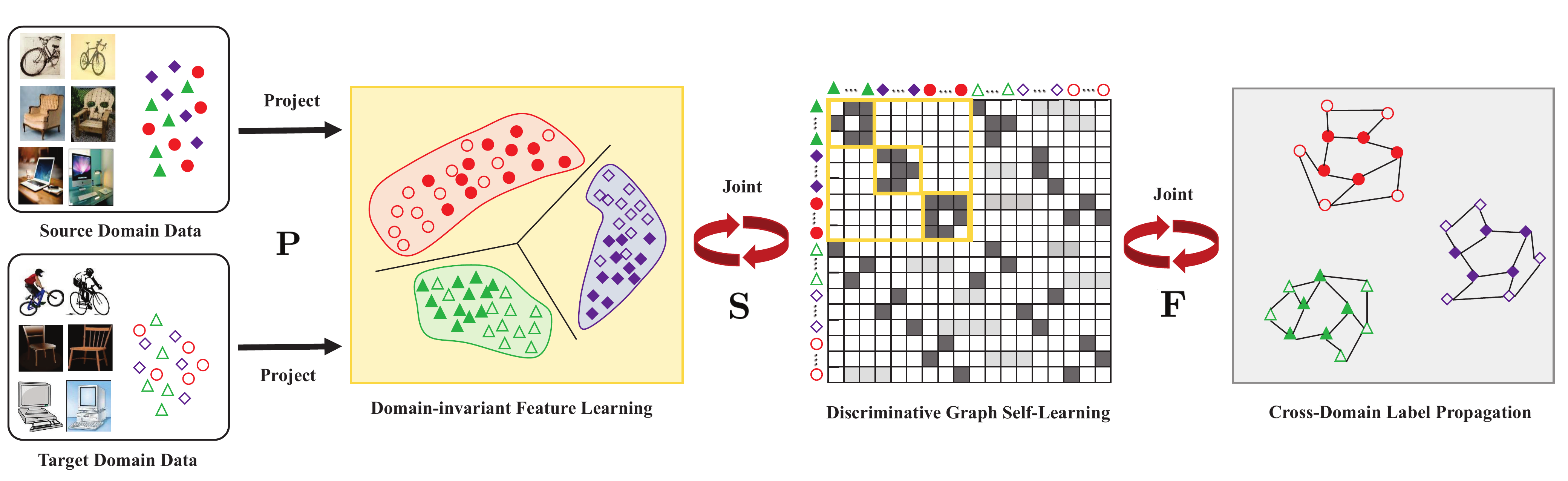}
	\caption{Flowchart of our proposed CDGS. We integrate domain-invariant feature learning, adaptive graph learning and cross-domain label propagation into a unified optimization framework. Besides, in order to construct a high-quality affinity matrix in our CDGS, we further propose a discriminative graph self-learning strategy. To be specific, instead of predefining the similarity metric, our proposal could adaptively assign neighbors for each sample according to the local distance in the projected feature space. To fully explore the discriminative information contained in well-labeled source data and pseudo-labeled target data, we further impose block diagonal structure constraint on source data and label smoothness constraint on two domain data.}
	\label{Framework}
\end{figure*}

The goal of domain adaptation is to mitigate the distribution discrepancy between the source and target domains, such that the classifier could be applicable across two domains. To accomplish this, numerous works \cite{JDA,DICD,DICE,DTLC,PACET} have  devoted to learning a domain-invariant space where distribution discrepancy can be significantly reduced via minimizing a distance metric, {\it e.g.}, the widely used maximum mean discrepancy (MMD) \cite{MMD}. Along this line, JDA \cite{JDA} is a pioneering method, which aims to reduce the joint distribution shift between two domains by simultaneously aligning the marginal distribution and conditional distribution. Inheriting the core idea of minimizing joint distribution discrepancy, tremendous subsequent studies following JDA \cite{DICD,DICE,PACET}, focus on further reducing the conditional distribution discrepancy by improving the prediction accuracy of target pseudo-labels. Despite the brilliant achievements in the literature, most of them generally overlook the   underlying  data manifold structure in the process of inferring data labels on target domain, thus making the performance of domain adaptation far from satisfactory. 

More recently, to explore the data distribution structure,  several studies \cite{DICE,DGA-DA,LPJT,GAKT} innovatively propose to infer target pseudo-labels by cross-domain label propagation \cite{LP}.
Generally, these methods follow a multi-stage paradigm in each iteration: 1) projecting the source and target data into a domain-invariant common feature space; 2) constructing a affinity matrix by calculating the sample similarity in the projected space with a predefined metric, {\it e.g.}, the gaussian kernel similarity \cite{DGA-DA,LPJT}; 3) assigning pseudo-labels for target data via propagating the labels of  source  data  with the constructed affinity matrix. Although improved performance has been achieved by these methods, they still suffer from three crucial issues: 
\begin{itemize}
	\item \textbf{Issue 1} The domain-invariant feature learning, affinity matrix constructing and target labels inferring are separated into three independent stages. Thus, the correlation among these three parts could not be fully exploited.
	\item \textbf{Issue 2}  Constructing the affinity matrix with predefined metric may not capture the inherent similarity of samples in both domains, which might seriously affect the effectiveness of cross-domain label propagation.
	\item \textbf{Issue 3} During the construction of  affinity matrix, the discriminative information contained in the ground-truth labels of source data, as well as in the pseudo-labels of target data  is less explored. 
\end{itemize}

In this study, we propose a novel domain adaptation method called Cross-domain label propagation with Discriminative Graph Self-learning (CDGS) to remedy the above three issues. As illustrated in Fig. \ref{Framework}, to tackle the first issue, we propose to formulate the three parts of cross-domain label propagation into a unified optimization framework, which learns domain-invariant features, constructs affinity matrix and infers target labels simultaneously. In the unified framework, these three parts can assist each other  from an iterative optimization perspective. For the second issue, inspired by \cite{PCAN,AWSSL}, we resort to a graph self-learning strategy, which assigns adaptive neighbors for each sample according to the local distance in the projected feature space. In such way, the underlying data manifold structure of two domains could be captured more effectively. 
To handle the third issue, for well-annotated source data,  we enforce the learned connected subgraph to have a block diagonal structure, which means that only source samples within the same category are allowed to be connected, while the connection weight of source samples between different categories is forcibly set to 0. In this manner, the discriminative information of source data can be exploited to the maximum extent. Beyond that,  inspired by \cite{AWSSL,USSL}, we further impose the label smoothness constraint during the graph self-learning, 
such that the weakly supervised information contained in target pseudo-labels  can be well inserted into the adaptive graph.


It is noteworthy that, except for  unsupervised domain adaptation (UDA), our CDGS could be readily extended to the semi-supervised domain adaptation (SDA) scenario   where  some labeled target samples are available. Interestingly, the extended SDA model could be solved with the same  algorithm as UDA.
To sum up, we list our contributions in fourfolds:
\begin{enumerate}
	\item We propose a novel cross-domain label propagation method for domain adaptation named CDGS, which integrates domain-invariant feature learning, affinity matrix constructing and target labels inferring into a unified optimization framework. Through the joint optimization, the three parts could boost each other and thus more effective knowledge transfer can be achieved.
	\item To construct a high-quality affinity matrix in CDGS, we propose a discriminative graph self-learning strategy, which can  not only adaptively capture the local connectivity structure of data from two domains  but also effectively explore the  discriminative information.
	\item An efficient optimization algorithm is designed to solve the objective function of our CDGS. In addition to UDA, we further extend  CDGS to the semi-supervised scenario in a direct but effective way and solve the extended  model with the identical optimization algorithm. 
	\item Extensive experiments on six standard datasets verify that the proposed CDGS  can consistently outperform the state-of-the-art methods in both UDA and SDA settings.
\end{enumerate}

The rest of this paper is organized as follows. Section \ref{Related work} provides a brief review on related domain adaptation and label propagation methods. Section \ref{Proposed Method} introduces the proposed CDGS approach, the optimization algorithm, the computational complexity  and the extension to SDA. Extensive experimental analysis is presented in Section \ref{experiments}. Finally, this paper is summarized   in Section \ref{conclusion}.

\section{Related Work \label{Related work}}
In this section, we review the related works in terms of domain adaptation and label propagation, and highlight the difference between the previous works and our proposal.

\subsection{Domain Adaptation}
Domain adaptation aims to leverage the knowledge from a well-labeled source domain to an unlabeled but related target domain. In general, domain adaptation can be grouped as UDA and SDA. In UDA, no labeled target samples are available. While in SDA, the target domain contains few labeled samples.

Generally, existing UDA methods can be roughly divided into three categories: instance reweighting \cite{CSA, TJM}, classifier adaptation \cite{ARTL, MEDA} and feature adaptation \cite{TCA,JDA,DICD,DICE} methods. Instance reweighting methods assign source samples with different weights to reduce the distribution shift between two domains. Classifier adaptation methods adapt the classifier trained on source data to target data. Feature adaptation methods seek a common feature space \cite{JDA} or latent intermediate subspaces \cite{GFK} to make the two domains have similar distributions. The proposed CDGS falls into the former line of feature adaptation methods, thus we focus on reviewing the works related to it. Among existing works, TCA \cite{TCA} proposes to align marginal distribution between two domains with MMD metric for the first time. Following this idea, JDA \cite{JDA} further considers the conditional distribution, such that the joint distribution alignment can be achieved. To boost the classification performance, several subsequent works propose to employ the discriminative information by encouraging intra-class compactness and inter-class dispersion \cite{DICD} simultaneously or promoting domain-irrelevant class clustering \cite{DICE}. To refine the target pseudo-labels to further mitigate the conditional distribution discrepancy, several recent works attempt to exploit the geometric structure underlying data manifold by assigning target pseudo-labels via cross-domain label propagation \cite{DICE,DGA-DA,LPJT,GAKT} or performing label propagation just on target domain \cite{DTLC,LSC}, and promising performance have been achieved by them. 

Our CDGS also employs cross-domain label propagation strategy to assign target pseudo-labels. However, CDGS is significantly different from these methods. First, CDGS integrates domain-invariant feature learning, affinity matrix constructing and target labels inferring into a unified optimization formulation while \cite{DICE,DGA-DA,LPJT} separate the three parts into independent stages, and \cite{GAKT} only combines the domain-invariant feature learning and target labels inferring. Through the joint optimization in our CDGS, the three parts could benefit from each other to yield a superior performance. Second, CDGS presents a novel self-learning strategy to construct a discriminative graph. Specifically, the neighbors of each sample are adaptively assigned according to the local distance, which is calculated based on the projected features and label information of source and target data. Besides, only source samples within the same class are enforced to be connected to exploit the source discriminative information. Thus, the discriminative graph can not only faithfully capture the inherent local connectivity structure of samples but also effectively explore the  discriminative information contained in source ground-truth labels and target  pseudo-labels, which is beneficial to effective target pseudo-labels assignment.

In the past few years, deep domain adaptation methods have attracted considerable interest and different strategies have been proposed to align deep features. For example, DAN \cite{DAN} exploits the multikernel MMD to reduce the marginal distribution discrepancy in the reproducing kernel Hilbert space (RKHS). Based on this framework, JAN \cite{JAN} proposes to align the joint distribution between two domains. To capture the fine-grained information, DSAN \cite{DSAN} further aligns the relevant subdomain distributions within the same category in two domains based on a local MMD. Different from them, DANN \cite{DANN} tries to learn domain agnostic feature representations with adversarial learning. Later, MADA \cite{MADA} trains a class-wise domain discriminator for each class. To enhance positive transfer and relieve negative transfer, Wang \emph{et al}. \cite{Wang2020} introduced a self-adaptive re-weighted adversarial approach to promote domain alignment in terms of conditional distribution. However, these deep methods may confront the challenges of long training time and massive resource consumption while CDGS is faster and can achieve excellent performance by just using off-the-shelf deep features.

Many methods have also been developed for SDA \cite{MMDT,CDLS,OBTL}. For instance, MMDT \cite{MMDT} learns the transformation matrix and classifier parameters jointly by making samples within the same class have high similarity. CDLS \cite{CDLS} aligns the conditional distribution by selecting representative landmarks. OBTL \cite{OBTL} is a Bayesian transfer learning framework, which relates the two domains by joint prior density. The proposed CDGS can be readily extended to SDA. Specifically, we take the labeled and unlabeled target data as a whole. In such case, we can estimate target class means more accurately, which can result in more accurate conditional distribution alignment. Besides, as a common strategy in semi-supervised learning, reliable connections between labeled and unlabeled data are built by discriminative graph self-learning, thus the knowledge from labeled samples can be propagated to the unlabeled ones. Moreover, the resulting optimization problem has the same formula as that of the unsupervised setting, thus they can be solved with the same optimization algorithm.
\subsection{Label Propagation}
The goal of label propagation is to propagate the label information of limited labeled samples to amounts of unlabeled samples through graph.  In the graph, a vertex represents a sample and the weight of the edge between two vertexes measures the similarity of the corresponding samples.

GFHF \cite{GFHF} and LGC \cite{LGC} are two classical methods. Both of them first use the gaussian kernel similarity to build the affinity matrix and then utilize label propagation to predict the unknown labels via gaussian fields and harmonic function, or the local and global consistency. However, they can not exploit the relationship of the affinity matrix and label information of samples due to the two separated stages. To overcome this limitation, STSSL \cite{USSL} integrates the affinity matrix constructing and the unknown labels inferring into one unified optimization framework to exploit the correlation between them. Following this idea, AWSSL \cite{AWSSL} futher proposes to adaptively assign the neighbors of each sample and effectively extract robust features by auto-weighting feature selection.

There are several classifier adaptation methods, which borrow the advantages of cross-domain label propagation to assign target pseudo-labels, {\it e.g.}, ARTL \cite{ARTL} and MEDA \cite{MEDA}. ARTL is also a unified framework, which learns an adaptive classifier by jointly optimizing the source structural risk, joint distribution alignment and manifold regulation, which is relevant to our CDGS. However, CDGS differs from ARTL in three aspects. First, ARTL learns the classifier with the original features while CDGS conducts subspace learning, which is more flexible and effective. Second, CDGS learns domain-invariant features, constructs affinity matrix and infers target labels jointly to fully exploit the relationship among them. Third, CDGS and ARTL use different strategies to construct the affinity matrix. Specifically, CDGS introduces a self-learning strategy to capture the intrinsic similarity of samples as well as effectively explore the label information of source and target data. By contrast, ARTL just utilizes the predefined metric to calculate the similarity for all samples.

\section{Proposed Method \label{Proposed Method}}
In this section, the key notations throughout this paper are first introduced. Then, we describe the details of the proposed CDGS. Next, we design an iterative algorithm to solve the optimization problem and provide the computational complexity analysis. Finally, we extend our method to SDA.
\subsection{Notations}
In UDA, the labeled source data $\mathcal{D}_s = \{\mathbf{X}_s,\mathbf{Y}_s\} = \{(\mathbf{x}_{si},y_{si})\}_{i=1}^{n_s}$ and unlabeled target data $\mathcal{D}_t = \{\mathbf{X}_t\} = \{\mathbf{x}_{tj}\}_{j=1}^{n_t}$ are given, where $\mathbf{x}_{si} \in \mathbb{R}^{m}$ is a source sample ($y_{si} \in \mathbb{R}$ is its label), $\mathbf{x}_{tj} \in \mathbb{R}^{m}$ is a target sample,  $n_s$ and $n_t$ represent the number of source and target samples. The entire data matrix is denoted as $\mathbf{X} =[ \mathbf{X}_s,\mathbf{X}_t] = \{\mathbf{x}_{i}\}_{i=1}^{n}$, where $n = n_s + n_t$.
For clarity, the key notations throughout this paper and their descriptions are summarized in Table \ref{notations}.

\begin{table}[]
	\centering
	\caption{Frequently Used Notations and Their Descriptions}
	\label{notations}
	\begin{tabular}{ccc}
		\toprule
		Notation & \quad Description\\
		\midrule
		$\mathbf{X}_s/\mathbf{X}_t/\mathbf{X}$&  \quad source/target/all data\\
		$n_s/n_t/n$&  \quad number of source/target/all data matrix\\
		$\mathbf{S}/\mathbf{L}$&  \quad affinity matrix/Laplacian matrix for all data\\
		$\mathbf{P}$& \quad  projection matrix\\
		$\mathbf{F}$& \quad  label matrix for all data\\
		$\mathbf{H}$& \quad  centering matrix\\
		$\mathbf{I}_d$ & \quad identity matrix with dimension $d$\\
		$m/d$ & \quad dimension of original/projected features\\
		$C$&  \quad number of shared classes\\
		$\mathbf{1}_{p \times q}$ &  \qquad a matrix of size $p \times q$ with all elements as $1$ \\
		$\mathbf{1}_{p}$ & \quad a column vector of size $p$ with all elements as $1$ \\
		\bottomrule
	\end{tabular}
\end{table}

\subsection{Problem Formulation \label{Problem Formulation}}
In this paper, we propose the CDGS framework to address domain adaptation problem, which integrates domain-invariant feature learning, affinity matrix constructing and target labels inferring into a unified optimization objective. The overall framework of our CDGS can be formulated as:
\begin{equation}
\label{framework}
\min_{\mathbf{P}, \mathbf{S}, \mathbf{F}}\Omega(\mathbf{P}, \mathbf{X}) + \alpha \Theta(\mathbf{P}, \mathbf{S}, \mathbf{X})+ \beta \Psi(\mathbf{F},\mathbf{S}) + \gamma\Phi(\mathbf{P})
\end{equation}
where $\mathbf{P} \in \mathbb{R}^{m \times d}$ denotes the projection matrix,  $\mathbf{F} \in \mathbb{R}^{n \times C}$ is the label matrix for all data and $\mathbf{S} \in \mathbb{R}^{n \times n}$ represents the affinity matrix. $\Omega(\mathbf{P}, \mathbf{X})$ is employed to learn  domain-invariant features. $\Theta(\mathbf{P}, \mathbf{S}, \mathbf{X})$ is utilized to adaptively construct the affinity matrix with the projected features. $\Psi(\mathbf{F},\mathbf{S})$ is used to infer the target labels by cross-domain label propagation. $\Phi(\mathbf{P})$ is the regularization term for the projection matrix to avoid overfitting. $\alpha$, $\beta$ and $\gamma$ are hyperparameters to balance the importance of different parts. As we can see, by integrating the three parts into the joint optimization objective, they could well communicate with each other to achieve more effective knowledge transfer. Next, more details about the three parts are presented.

\subsubsection{Domain-invariant Feature Learning} When $\mathbf{X}_s$ and $\mathbf{X}_t$ are drawn from different feature distributions, it is crucial to reduce the distribution discrepancy between two domains, such that the classifier trained on source data can be directly applied to target domain. To measure the distribution discrepancy, numerous metrics have been proposed. Among them, MMD \cite{MMD} is probably the most widely used one. In the projected space, the MMD distance between two domains can be calculated as the distance between the sample means of the source and target data \cite{DICD}. Considering the large distribution discrepancy across domains, we minimize the marginal distribution distance and the conditional distribution distance simultaneously, and denote them by $\mathcal{L}_{mmd}^{m}$ and $\mathcal{L}_{mmd}^{c}$, respectively. With the MMD metric, marginal distribution distance can be stated as:
\begin{equation}
\label{marginal}
\begin{aligned}
\mathcal{L}_{mmd}^{m}
&=\|\frac{1}{n_s}\sum\limits_{i=1}^{n_s}\mathbf{P}^\mathrm{T}\mathbf{x}_{si}-\frac{1}{n_t}\sum\limits_{j=1}^{n_t}\mathbf{P}^\mathrm{T}\mathbf{x}_{tj}\|_2^2 \\
&=\mathrm{tr}(\mathbf{P}^\mathrm{T}\mathbf{X}\mathbf{M}_0\mathbf{X}^\mathrm{T}\mathbf{P})
\end{aligned}
\end{equation}
where $\mathrm{tr}(\cdot)$ is the trace operator, and $\mathbf{M}_0 \in \mathbb{R}^{n \times n}$ represents the marginal MMD matrix calculated as: 
\begin{equation}
\label{M0}
\begin{aligned}
\mathbf{M}_0 = 
\begin{bmatrix}

\frac{1}{n_s^2}\mathbf{1}_{n_s \times n_s}& -\frac{1}{n_sn_t}\mathbf{1}_{n_s \times n_t} \\

-\frac{1}{n_sn_t}\mathbf{1}_{n_t \times n_s} & \frac{1}{n_t^2}\mathbf{1}_{n_t \times n_t}
\end{bmatrix}
\end{aligned}
\end{equation}

The calculation of conditional MMD distance requires to get the labels of target samples, which is generally infeasible in domain adaptation task. To remedy this issue, we employ the target pseudo-labels instead of the unavailable true labels to compute the conditional distribution distance  as follows:
\begin{equation}
\label{conditional}
\begin{aligned}
\mathcal{L}_{mmd}^{c}
&=\sum\limits_{c=1}^{C}\|\frac{1}{n_s^c}\sum\limits_{\mathbf{x}_{si} \in \mathbf{X}_s^c}\mathbf{P}^\mathrm{T}\mathbf{x}_{si}-\frac{1}{n_t^c}\sum\limits_{\mathbf{x}_{tj} \in \mathbf{X}_t^c}\mathbf{P}^\mathrm{T}\mathbf{x}_{tj}\|_2^2 \\
&=\mathrm{tr}(\mathbf{P}^\mathrm{T}\mathbf{X}(\sum\nolimits_{c=1}^C\mathbf{M}_c)\mathbf{X}^\mathrm{T}\mathbf{P})
\end{aligned}
\end{equation}
where $C$ is the number of classes, $\mathbf{M}_c \in \mathbb{R}^{n \times n}$ is conditional MMD matrix defined as:
\begin{equation}
\label{MMD matrix}
(\mathbf{M}_c)_{ij} =
\begin{cases}
\frac{1}{n_s^cn_s^c}, &\mathrm{if} \ \mathbf{x}_i,\mathbf{x}_j \in \mathbf{X}_s^c;\\
\frac{1}{n_t^cn_t^c}, &\mathrm{if} \ \mathbf{x}_i,\mathbf{x}_j \in \mathbf{X}_t^c;\\
\frac{-1}{n_s^cn_t^c}, &\mathrm{if} \ \mathbf{x}_i \in \mathbf{X}_s^c \wedge \mathbf{x}_{j} \in \mathbf{X}_t^c;\\
\frac{-1}{n_s^cn_t^c}, &\mathrm{if} \ \mathbf{x}_j \in \mathbf{X}_s^c \wedge \mathbf{x}_{i} \in \mathbf{X}_t^c;\\
0, &\text{otherwise}
\end{cases}
\end{equation}
$\mathbf{X}_s^c$ represents all source samples in class $c$, and $n_s^c$ is the corresponding number of samples. Similar definitions can be applied for target samples according to the pseudo-labels. Denote $\mathbf{M} = \sum\nolimits_{c=0}^C\mathbf{M}_c$, then we have the following formula:
\begin{equation}
\label{MMD}
\begin{aligned}
\Omega(\mathbf{P}, \mathbf{X}) = \mathrm{tr}(\mathbf{P}^\mathrm{T}\mathbf{XMX}^\mathrm{T}\mathbf{P})
\end{aligned}
\end{equation}

\subsubsection{Graph Self-Learning with Source Domain Discriminative Structure Preserving} Obviously, the quality of the affinity matrix is crucial to the performance of cross-domain label propagation. Most of previous works \cite{DICE, DGA-DA,LSC} use the same strategy to construct it, which calculates the similarity for all samples with predefined similarity metric, {\it e.g.}, the heatkenel similarity \cite{DICE,DGA-DA}. This strategy may not capture the inherent similarity of samples, thus hinders the correctness of cross-domain label propagation and results in serious misclassification for target data. The wrong pseudo-labels will further mislead the conditional distribution alignment in the next iteration, which ultimately results in significant performance degradation. To tackle this issue, inspired by several recent works \cite{PCAN, AWSSL}, we adopt a self-learning strategy, which constructs the affinity matrix by assigning the adaptive neighbors for each sample according to the local distance in the projected space. In light of this, the optimization objective of graph self-learning can be stated as follows:
\begin{equation}
\label{asm}
\begin{aligned}
&\min_{\mathbf{S}}\sum\nolimits_{i=1}^n((\sum\nolimits_{j=1}^n\|\mathbf{z}_i-\mathbf{z}_j\|_2^2S_{ij} + \lambda_i\|\mathbf{S}_{i,:}\|_2^2)\\  
& =\min_{\mathbf{S}}\mathrm{tr}(\mathbf{P}^\mathrm{T}\mathbf{XL}\mathbf{X}^\mathrm{T}\mathbf{P}) + \|\mathbf{\Lambda}\mathbf{S}\|_F^2\\
&s.t. \quad \mathbf{S}\mathbf{1}_n = \mathbf{1}_n, \ 0 \leq S_{ij} \leq 1
\end{aligned}
\end{equation}
where $\mathbf{z}_i = \mathbf{P}^\mathrm{T}\mathbf{x}_i$ is the projection of sample $\mathbf{x}_i$, $\mathbf{S}_{i,:}$ represents the $i$-th row of $\mathbf{S}$, and $\mathbf{\Lambda} = \mathrm{diag}(\sqrt{\lambda_1}, \sqrt{\lambda_2}, ... , \sqrt{\lambda_n})$. $\mathbf{L}$ is the graph Laplacian matrix calculated as $\mathbf{L} = \mathbf{D} - \mathbf{S}$, and $\mathbf{D}$ is a diagonal matrix with the $i$-th element $D_{ii} = \sum\nolimits_{j=1}^nS_{ij}$. An $F$-norm regulation term is imposed on the $i$-th ($i = 1,2,\ldots,n$) row of $\mathbf{S}$ and the corresponding regularization term is $\lambda_i$, which can be determined automatically and will be elaborated  in Section \ref{Optimization Procedure}.  Then, we can obtain the following formula for $\Theta(\mathbf{P}, \mathbf{S}, \mathbf{X})$:
\begin{equation}
\label{GFHF3}
\Theta(\mathbf{P}, \mathbf{S}, \mathbf{X}) = \mathrm{tr}(\mathbf{P}^\mathrm{T}\mathbf{XL}\mathbf{X}^\mathrm{T}\mathbf{P}) + \|\mathbf{\Lambda}\mathbf{S}\|_F^2
\end{equation}

In addition, several previous works \cite{DICD, DICE,DGA-DA} have shown that the performance of domain adaptation can be significantly enhanced if the discriminative information of source data is exploited. To this end, we adopt an intuitive strategy for labeled source data that only  the samples belonging to the same category are allowed to be connected. In such case, each source sample could be connected with two parts, one of which is the source samples within the identical class and the other is all target samples. For simplicity, we fix the probability as $\delta$ and $1-\delta$ for these two parts, respectively. That is, when $i \leq n_s$, we have $\sum\nolimits_{j=1}^{n_s}S_{ij} = \delta$ and $\sum\nolimits_{j=n_s+1}^{n_s+n_t}S_{ij} = 1-\delta$, where $\delta \in [0, 1]$ is a hyperparameter to control the partition of probability. In this way, the learned adaptive discriminative graph owns the following structure:
\begin{equation}
	\mathbf{S} = \left[
	\begin{array}{cc}
	\begin{array}{c|c}
	\overbrace{
		\begin{array}{cccc}
		\mathbf{S}_{ss}^{(1)} & \mathbf{0} & \cdots & \mathbf{0}\\
		\mathbf{0} & \mathbf{S}_{ss}^{(2)} & \cdots & \mathbf{0}\\
		\vdots & \vdots & \ddots & \vdots\\
		\mathbf{0} & \mathbf{0} & \cdots & \mathbf{S}_{ss}^{(C)}
		\end{array}
	}^{\delta}
	& 
	\overbrace{
		\begin{array}{ccc}
		& & \\
		& \\
		& \mathbf{S}_{st}& \\
		& & \\
		& &
		\end{array}
	}^{1-\delta} 
	\end{array} \\ \hline
	\begin{array}{ccccccc}
	&&&&&&\\
	&&&&\mathbf{S}_{ta}&&\\
	&&&&&&\\
	\end{array}
	\end{array}
	\right]
\end{equation}
where 
$\mathbf{S}_{ss}^{(c)}$ represents the affinity matrix of all source samples in class $c$, $\mathbf{S}_{ss}$ is the connected subgraph with block diagonal structure for all source samples,  $\mathbf{S}_{st}$ is the connected subgraph between source and target samples, and $\mathbf{S}_{ta}$ denotes the connected subgraph between target samples and all samples. Then, considering the above constraints, the objective of graph self-learning with source domain discriminative structure preserving term $\Theta(\mathbf{P}, \mathbf{S}, \mathbf{X})$ can be formulated as: 
\begin{equation}
\label{asm1}
\begin{aligned}
&\min_{\mathbf{S}}\sum\nolimits_{i=1}^n((\sum\nolimits_{j=1}^n\|\mathbf{z}_i-\mathbf{z}_j\|_2^2S_{ij} +\lambda_i\|\mathbf{S}_{i,:}\|_2^2)\\  
& =\min_{\mathbf{S}}\mathrm{tr}(\mathbf{P}^\mathrm{T}\mathbf{XL}\mathbf{X}^\mathrm{T}\mathbf{P}) + \|\mathbf{\Lambda}\mathbf{S}\|_F^2\\
&s.t. \ \mathbf{S}\mathbf{1}_n = \mathbf{1}_n, \ 0 \leq S_{ij} \leq 1, \ \sum\limits_{j=1}^{n_s}S_{ij} = \delta, \ i \leq n_s, \\
& \qquad \ S_{ij} = 0, \ i,j \leq n_s\wedge y_{si} \ne y_{sj}
\end{aligned}
\end{equation}

\subsubsection{Cross-Domain Label Propagation with Self-Learned Graph}
The main assumption of label propagation \cite{LP} is that the adjacent  points tend to  similar labels. Thus, assigning the target pseudo-labels by cross-domain label propagation could effectively exploit the geometric structure underlying the source and target data manifolds, which can improve the accuracy of target pseudo-label prediction  and further benefit the subsequent domain-invariant feature learning. To achieve this goal, numerous current label propagation algorithms can be our candidates. For the sake of simplicity, we employ the classical GFHF algorithm \cite{GFHF}. Suppose the source label matrix is $\mathbf{F}_s \in \mathbb{R}^{n_s \times C}$, where the ($i$, $j$)-th element is 1 if $y_{si} = j$, and 0 otherwise. Denote $\mathbf{F} = [\mathbf{F}_l;\mathbf{F}_t]$, where $\mathbf{F}_l = \mathbf{F}_s$, and $\mathbf{F}_t \in \mathbb{R}^{n_t \times C}$ is the inferred target label matrix. Then, given the self-learned graph $\mathbf{S}$, the optimization problem of the GFHF algorithm can be formulated as:
\begin{equation}
\label{GFHF}
\min_{\mathbf{F}}\sum\limits_{i,j = 1}^n\|\mathbf{F}_i - \mathbf{F}_j\|_2^2S_{ij} = \min_{\mathbf{F}}\mathrm{tr}(\mathbf{F}^\mathrm{T}\mathbf{L}\mathbf{F}) \ s.t. \ \mathbf{F}_l = \mathbf{F}_s
\end{equation}
Thus, we can obtain the following formula for $\Psi(\mathbf{F},\mathbf{S})$:
\begin{equation}
\label{GFHF2}
\Psi(\mathbf{F},\mathbf{S}) = \mathrm{tr}(\mathbf{F}^\mathrm{T}\mathbf{L}\mathbf{F})
\end{equation}

\subsubsection{The Final Form for Discriminative Graph Self-Learning} Actually, the pseudo-labels of target data has been proven to be able to provide useful ``weakly" supervised information to yield a better connectivity graph \cite{AWSSL,USSL}. In our CDGS, 
we further impose {\it label smoothness constraint} ({\it i.e.}, Eq.(\ref{GFHF2})) during  graph self-learning to effectively explore the  target pseudo-label information.
Thus, our final optimization objective of discriminative graph self-learning can be formulated as:
\begin{equation}
\label{AMC}
\begin{aligned}
&\min_{\mathbf{S}}\sum\nolimits_{i=1}^n((\sum\nolimits_{j=1}^n\|\mathbf{z}_i-\mathbf{z}_j\|_2^2S_{ij} +\beta \|\mathbf{F}_i - \mathbf{F}_j\|_2^2S_{ij})\\
&\qquad \qquad +\lambda_i\|\mathbf{S}_{i,:}\|_2^2)\\  
& \quad =\min_{\mathbf{S}}\mathrm{tr}(\mathbf{P}^\mathrm{T}\mathbf{XL}\mathbf{X}^\mathrm{T}\mathbf{P}) + \|\mathbf{\Lambda}\mathbf{S}\|_F^2 + \beta \mathrm{tr}(\mathbf{F}^\mathrm{T}\mathbf{L}\mathbf{F})\\
&s.t. \ \mathbf{S}\mathbf{1}_n = \mathbf{1}_n, \ 0 \leq S_{ij} \leq 1, \ \sum\limits_{j=1}^{n_s}S_{ij} = \delta, \ i \leq n_s, \\
& \qquad \ S_{ij} = 0, \ i,j \leq n_s\wedge y_{si} \ne y_{sj}
\end{aligned}
\end{equation}

To avoid overfitting and improve the generalization capacity, we further impose an $F$-norm regularization term on $\mathbf{P}$, then we have:
\begin{equation}
\label{reg}
\Phi(\mathbf{P}) = \|\mathbf{P}\|_F^2
\end{equation}

Finally, by combining Eq. (\ref{MMD}), Eq. (\ref{GFHF}),  Eq. (\ref{AMC}) and Eq. (\ref{reg}), we obtain the final formulation of our CDGS:
\begin{equation}
\label{GFHF-obj}
\begin{aligned}
&\min_{\mathbf{P},\mathbf{S},\mathbf{F}}\mathrm{tr}(\mathbf{P}^\mathrm{T}\mathbf{XMX}^\mathrm{T}\mathbf{P}) + \alpha(\mathrm{tr}(\mathbf{P}^\mathrm{T}\mathbf{X}\mathbf{L}\mathbf{X}^\mathrm{T}\mathbf{P}) \\
& \qquad \ + \|\mathbf{\Lambda S}\|_F^2) + \beta \mathrm{tr}(\mathbf{F}^\mathrm{T}\mathbf{L}\mathbf{F}) + \gamma\|\mathbf{P}\|_F^2\\
&\ \ s.t. \ \mathbf{P}^\mathrm{T}\mathbf{XHX}^\mathrm{T}\mathbf{P} = \mathbf{I}_d, \ \mathbf{S}\mathbf{1}_n = \mathbf{1}_n, 0 \leq S_{ij} \leq 1,\\
& \qquad \ \mathbf{F}_l = \mathbf{F}_s, \ \sum\nolimits_{j=1}^{n_s}S_{ij} = \delta, \ i \leq n_s, \\
& \qquad \ S_{ij} = 0, \ i,j \leq n_s\wedge y_{si} \ne y_{sj}
\end{aligned}
\end{equation}
where $\mathbf{H}$ is the centering matrix defined as $\mathbf{H} = \mathbf{I}_{n}- \frac{1}{n}\mathbf{1}_{n \times n}$. The first constraint is to maximize the variance of all data \cite{JDA} in the projected space, which is inspired by the principal component analysis. Similar to \cite{DICD}, $\mathbf{M}$ and $\mathbf{L}$ can be normalized into the same scale. Thus, we  set $\alpha = 1.0$ for all cases.

\subsection{Optimization Procedure \label{Optimization Procedure}}
In problem (\ref{GFHF-obj}), we need to optimize three variables  $\mathbf{P}$, $\mathbf{S}$, and $\mathbf{F}$. As it is not jointly convex with the three variables, we update each variable alternatively with the others fixed. To be specific, we solve each subproblem as follows.

\textbf{1. $\mathbf{P}$-Subproblem:} When we fix $\mathbf{S}$ and $\mathbf{F}$, the optimization problem (\ref{GFHF-obj}) becomes:
\begin{equation}
\begin{aligned}
&\min_{\mathbf{P}}\mathrm{tr}(\mathbf{P}^\mathrm{T}(\mathbf{X}\mathbf{M}\mathbf{X}^\mathrm{T} + \alpha \mathbf{X}\mathbf{L}\mathbf{X}^\mathrm{T} + \gamma \mathbf{I}_m)\mathbf{P}) \\
&\qquad s.t. \ \mathbf{P}^\mathrm{T}\mathbf{XHX}^\mathrm{T} = \mathbf{I}_d
\end{aligned}
\end{equation}
We employ the Lagrange techniques to solve it. The corresponding Lagrangian function can be formulated as:
\begin{equation}
\label{LF}
\begin{aligned}
L(\mathbf{P},\mathbf{\Theta}) = &\ \mathrm{tr}(\mathbf{P}^\mathrm{T}(\mathbf{X}\mathbf{M}\mathbf{X}^\mathrm{T} + \alpha \mathbf{X}\mathbf{L}\mathbf{X}^\mathrm{T} + \gamma \mathbf{I}_m)\mathbf{P}) \\
&+\mathrm{tr}((\mathbf{I}_d - \mathbf{P}^\mathrm{T}\mathbf{XHX}^\mathrm{T}\mathbf{P})\mathbf{\Pi})
\end{aligned}
\end{equation}
where $\mathbf{\Pi} = \mathrm{diag}(\pi_1, \pi_2, ..., \pi_d) \in \mathbb{R}^{d \times d}$ is a diagonal matrix and each element is a Lagrange Multiplier. By setting the gradient of (\ref{LF}) with respect to $\mathbf{P}$ to zero, we obtain:
\begin{equation}
\label{P-sub}
(\mathbf{X}\mathbf{M}\mathbf{X}^\mathrm{T} + \alpha \mathbf{X}\mathbf{L}\mathbf{X}^\mathrm{T} + \gamma \mathbf{I}_m)\mathbf{P} = \mathbf{XHX}^\mathrm{T}\mathbf{P}\mathbf{\Theta}
\end{equation}
Then the optimal solution can be obtained by computing the eigenvectors of (\ref{P-sub})
regarding to the $d$-smallest eigenvalues.

\textbf{2. $\mathbf{S}$-Subproblem:} When $\mathbf{P}$ and $\mathbf{F}$ are fixed, the optimization problem (\ref{GFHF-obj}) with regard to $\mathbf{S}$ is equal to problem (\ref{AMC}). Actually, problem (\ref{AMC}) can be divided into $n$ subproblems and each of them is formulated as:
\begin{equation}
\label{S-sub1}
\begin{aligned}
&\min_{\mathbf{S}_{i,:}}\sum\nolimits_{j=1}^n(\|\mathbf{z}_i-\mathbf{z}_j\|_2^2S_{ij}+\beta \|\mathbf{F}_i - \mathbf{F}_j\|_2^2S_{ij})\\
&\qquad +\lambda_i\|\mathbf{S}_{i,:}\|_2^2 \\
&s.t. \ \mathbf{S}_{i,:}\mathbf{1}_{n} = 1,\ 0 \leq S_{ij} \leq 1, \ \sum\nolimits_{j=1}^{n_s}S_{ij} = \delta, \ i \leq n_s \\
&\quad \ S_{ij} = 0, \ i,j \leq n_s\wedge y_{si} \ne y_{sj}
\end{aligned}
\end{equation}

\textbf{Case 1:}
First of all, we show how to obtain the optimal solution when $i \ \textgreater \ n_s$. We define $A_{ij} = \|\mathbf{z}_i-\mathbf{z}_j\|_2^2 + \beta \|\mathbf{F}_i - \mathbf{F}_j\|_2^2$, then the above problem can be reformulated as:
\begin{equation}
\label{S-sub2}
\min_{\mathbf{S}_{i,:}\mathbf{1}_{n} = 1,0 \leq S_{ij} \leq 1}\|\mathbf{S}_{i,:} + \frac {\mathbf{A}_{i,:}}{2\lambda_i}\|_2^2
\end{equation}

\begin{algorithm}[h]
	\SetAlgoLined
	\caption{CDGS Algorithm}
	\label{alg1}
	\KwIn{Source data $\{\mathbf{X}_s,\mathbf{Y}_s\}$;
		Target data $\{\mathbf{X}_t\}$; Initial affinity matrix $\mathbf{S}$;
		Hyper-parameters $\alpha = 1.0$, $k = 20$, $\delta = 0.8$, $\beta$, $\gamma$, $d$; Maximum iteration $T$ = 10.}
	\KwOut{Target pseudo-labels $\mathbf{\widehat{Y}}$.}
	\BlankLine
	$t$ = 1;\\
	\While { \rm{not converge} \textbf{and} $t$ $\leq T$}
	{
		// \textit{Projection matrix $\mathbf{P}$} \\
		Update $\mathbf{P}$ by solving (\ref{P-sub});\\
		// \textit{Affinity matrix $\mathbf{S}$} \\
		Update each row of $\mathbf{S}$ by (\ref{S-sub6}), (\ref{S-sub8}) and (\ref{S-sub10});\\
		// \textit{Target label matrix $\mathbf{F}_t$} \\
		Update $\mathbf{F}_t$ by (\ref{GFHF1});\\
		//\textit{Target pseudo-labels $\mathbf{\widehat{Y}}$} \\
		Update the target pseudo-labels $\mathbf{\widehat{Y}}$ by (\ref{pseudo label});\\
		$t$ = $t$ + 1;
	}
	\textbf{Return} Target pseudo-labels $\mathbf{\widehat{Y}}$.
\end{algorithm}

The corresponding Lagrangian function is
\begin{equation}
\label{S-sub3}
\min_{\mathbf{S}_{i,:}}\|\mathbf{S}_{i,:} + \frac {\mathbf{A}_{i,:}}{2\lambda_i}\|_2^2 - \mu(\mathbf{S}_{i,:}\mathbf{1}_{n} - 1) - \mathbf{S}_{i,:}\bm{\eta}^{\mathrm{T}}
\end{equation}
where $\mu$ and $\bm{\eta}$ are the Lagrangian multipliers. 
To utilize the local structure of data and relieve computation burden, we learn a sparse $\mathbf{S}_{i,:}$, i.e., each sample is only locally connected with its $k$-nearest neighbors. Based on the KKT condition, problem (\ref{S-sub3})  has a closed-form solution as follows:
\begin{equation}
\label{S-sub4}
S_{ij} = \mathrm{max}(z-\frac{A_{ij}}{2\lambda_i},0)
\end{equation}
where $z = \frac{1}{k} + \frac{1}{2k\lambda_i}\sum\nolimits_{j=1}^{k}\tilde{A}_{ij}$ and $\tilde{A}_{ij}$ is the entry of matrix $\tilde{\mathbf{A}}$, which is obtained by sorting the elements of each row of $\mathbf{A}$ from small to large. To ensure that each $\mathbf{S}_{i,:}$ has exactly $k$ nonzero elements, we could set $z - \tilde{A}_{i,k+1}/(2\lambda_i) = 0$, then we have:

\begin{equation}
\label{S_sub5}
\lambda_i = \frac{1}{2}(k\tilde{A}_{i,k+1} - \sum\nolimits_{j=1}^{k}\tilde{A}_{ij})
\end{equation}
Submitting Eq. (\ref{S_sub5}) into Eq. (\ref{S-sub4}), we can obtain:
\begin{equation}
\label{S-sub6}
S_{ij} = \mathrm{max}(\frac{\tilde{A}_{i,k + 1} - A_{ij}}{k\tilde{A}_{i,k+1} - \sum\nolimits_{j=1}^{k}\tilde{A}_{ij}},0)
\end{equation}

\textbf{Case 2:} When $i,j \leq n_s$, Eq.(\ref{S-sub1}) can be reformulated as:
\begin{equation}
\label{S-sub7}
\begin{aligned}
&\min_{\mathbf{S}_{i,:}}\sum\nolimits_{j=1}^{n_s}\|\mathbf{z}_i-\mathbf{z}_j\|_2^2S_{ij}+\beta \|\mathbf{F}_i - \mathbf{F}_j\|_2^2S_{ij}+\lambda_i S_{ij}^2\\
&s.t. \sum\nolimits_{j=1}^{n_s}S_{ij} = \delta, \ 0 \leq S_{ij} \leq 1, \ S_{ij} = 0, \ y_{si} \ne y_{sj}
\end{aligned}
\end{equation}
To satisfy the last constraint, we could set $A_{ij} = +\infty$ if $y_{si} \ne y_{sj}$. Similar to problem (\ref{S-sub2}), we can obtain the closed-form solution of problem (\ref{S-sub7}): 
\begin{equation}
\label{S-sub8}
S_{ij} = \delta\mathrm{max}(\frac{\tilde{A}_{i,k_1 + 1} - A_{ij}}{k_1\tilde{A}_{i,k_1+1} - \sum\nolimits_{j=1}^{k_1}\tilde{A}_{ij}},0)
\end{equation}
where $k_1 = \mathrm{min}(k,n_s^{y_{si}})$ as in practice, some classes may have very small-size samples.

\textbf{Case 3:}
When $i \leq n_s, j > n_s$, problem (\ref{S-sub1}) can be rewritten as:
\begin{equation}
\label{S-sub9}
\begin{aligned}
&\min_{\mathbf{S}_{i,:}}\sum\nolimits_{j=n_s+1}^n\|\mathbf{z}_i-\mathbf{z}_j\|_2^2S_{ij}+\beta \|\mathbf{F}_i - \mathbf{F}_j\|_2^2S_{ij} \\
& \qquad +\lambda_i S_{ij}^2\\
&s.t. \sum\nolimits_{j=n_s + 1}^{n}S_{ij} = 1-\delta,\ 0 \leq S_{ij} \leq 1
\end{aligned}
\end{equation}

Similarly, the closed-form solution of problem (\ref{S-sub9}) is:
\begin{equation}
\label{S-sub10}
S_{ij} = (1-\delta)\mathrm{max}(\frac{\tilde{A}_{i,k + 1} - A_{ij}}{k\tilde{A}_{i,k+1} - \sum\nolimits_{j=1}^{k}\tilde{A}_{ij}},0)
\end{equation}

\textbf{3. $\mathbf{F}$-Subproblem:} With fixed $\mathbf{P}$ and $\mathbf{S}$, the optimization problem (\ref{GFHF-obj}) with respect to $\mathbf{F}$ is equal to solve problem (\ref{GFHF}). According to \cite{GFHF}, we only need to update $\mathbf{F}_t$. Split $\mathbf{L}$ into four blocks: $\mathbf{L} = 
\begin{bmatrix} 
\mathbf{L}_{ss} & \mathbf{L}_{st} \\ \mathbf{L}_{ts} & \mathbf{L}_{tt} 
\end{bmatrix}
$, where $\mathbf{L}_{ss} \in \mathbb{R}^{n_s \times n_s}$, $\mathbf{L}_{st} \in \mathbb{R}^{n_s \times n_t}$, $\mathbf{L}_{ts} \in \mathbb{R}^{n_t \times n_s}$ and $\mathbf{L}_{tt} \in \mathbb{R}^{n_t \times n_t}$, and then, the optimal solution of problem (\ref{GFHF}) is:
\begin{equation}
\label{GFHF1}
\mathbf{F}_t = - \mathbf{L}_{tt}^{-1}\mathbf{L}_{ts}\mathbf{F}_s
\end{equation}

Eventually, the target pseudo-labels can be obtained based on the following decision function:
\begin{equation}
\label{pseudo label}
\widehat{y}_{ti} = \mathrm{argmax}_j\ (\mathbf{F}_t)_{ij}
\end{equation}

The affinity matrix $\mathbf{S}$ is initialized according to (\ref{S-sub8}) in the original feature space. We summarize the detailed optimization steps of the proposed CDGS in Algorithm \ref{alg1}.

\subsection{Computational Complexity Analysis}
To find the optimal solutions for the optimization Algorithm \ref{alg1}, we need to solve three subproblems. The complexity of each subproblem in each iteration is induced as follows: First, constructing and solving the eigen-decomposition problem (\ref{P-sub}) for $\mathbf{P}$-subproblem costs $\mathcal{O}(n^2m + dm^2)$; Then, updating the affinity matrix $\mathbf{S}$ needs a time cost of $\mathcal{O}(n^2\mathrm{log}(n))$; Finally, the complexity of obtaining the target estimated label matrix $\mathbf{F}_t$ and the pseudo-labels $\mathbf{\widehat{Y}}$ is $\mathcal{O}(n_t^3)$. Thus, the overall computational complexity of our proposal is $\mathcal{O}(Tn^2m+Tdm^2+Tn^2\mathrm{log}(n)+Tn_t^3)$, where $T$ is the number of iterations.

\subsection{Extension to Semi-supervised Domain Adaptation}
We denote the target data as $\mathbf{X}_t = [\mathbf{X}_l, \mathbf{X}_u]$, where $\mathbf{X}_l = \{\mathbf{x}_{li}\}_{i=1}^{n_l}$ is the labeled data and $\mathbf{X}_u = \{\mathbf{x}_{uj}\}_{j=1}^{n_u}$ is the unlabeled data. Then, by submitting $\mathbf{X}_s$ and  $\mathbf{X}_{t}$ into Eq. (\ref{GFHF-obj}), the semi-supervised extension for our CDGS can be stated as:
\begin{equation}
\label{semi-obj}
\begin{aligned}
&\min_{\mathbf{P},\mathbf{S},\mathbf{F}}\mathrm{tr}(\mathbf{P}^\mathrm{T}\mathbf{XMX}^\mathrm{T}\mathbf{P}) + \alpha(\mathrm{tr}(\mathbf{P}^\mathrm{T}\mathbf{X}\mathbf{L}\mathbf{X}^\mathrm{T}\mathbf{P}) \\
& \qquad \ + \|\mathbf{\Lambda S}\|_F^2) + \beta \mathrm{tr}(\mathbf{F}^\mathrm{T}\mathbf{L}\mathbf{F}) + \gamma\|\mathbf{P}\|_F^2\\
&\ \ s.t. \ \mathbf{P}^\mathrm{T}\mathbf{XHX}^\mathrm{T}\mathbf{P} = \mathbf{I}_d, \ \mathbf{S}\mathbf{1}_n = \mathbf{1}_n, 0 \leq S_{ij} \leq 1,\\
& \qquad \ \mathbf{F}_l = \mathbf{F}_{s}, \ \sum\nolimits_{j=1}^{n_s}S_{ij} = \delta, \ i \leq n_s, \\
& \qquad \ S_{ij} = 0, \ i,j \leq n_s\wedge y_{si} \ne y_{sj}
\end{aligned}
\end{equation}
where $n = n_s + n_l+n_u$. Obviously, Eq. (\ref{semi-obj}) owns the same formula with Eq. (\ref{GFHF-obj}), thus they can be solved with the identical algorithm.

Actually, our semi-supervised extension can be effective for the following two reasons: 1) The estimation of target class means is more accurate when some labeled target samples are available, which can promote to perform conditional distribution alignment more accurately; 2) Through Eq. (\ref{asm}), reliable connections between the labeled and unlabeled data are built, which can transfer the knowledge of labeled samples to the unlabeled ones via cross-domain label propagation.

\section{Experiments \label{experiments}}
In this section, we first describe the six benchmark datasets. Then, the details of experimental setup are shown. Next, we present the evaluation results of UDA, ablation study, parameter sensitivity and convergence analysis. Finally, the results for SDA are reported. The source code of this paper is available at https://drive.google.com/drive/folders/19Fqxxuf9MTcd-1em\\XstZE01G60JUyAst?usp=sharing.
\subsection{Datasets and Descriptions}
We adopt six benchmark datasets in our experiments, including Office31, Office-Caltech10, ImageNet-VOC2007, Office-Home, MNIST-USPS and PIE, which are widely used cross-domain object, digit and face datasets. Overall descriptions about these datasets are summarized in Table \ref{datasets}. We will introduce more details for each dataset as follows.

\emph{Office31} \cite{Office31} contains 4,110 images with 31 categories collected from three domains: Amazon (A), DSLR (D) and Webcam (W). Amazon images are downloaded from the online merchants. DSLR images are captured by a digital SLR camera while Webcam images are recorded by a web camera. Following \cite{CORAL}, we utilize the AlexNet-FC$_7$ features\footnote{ \url{https://github.com/VisionLearningGroup/CORAL/tree/master/dataset}} fine-tuned on the source domain.

\emph{Office-Caltech10} \cite{GFK} includes 2,533 images in 10 shared categories from the Office31 dataset and the Caltech256 (C) dataset, which is a widely used dataset for object recognition. Following \cite{GFK}, we exploit the SURF features\footnote{\url{http://boqinggong.info/assets/GFK.zip}}. Besides, the VGG-FC$_{6,7}$ features\footnote{ \url{https://sherath@bitbucket.org/sherath/ils.git}} provided by \cite{ILS} are used. 

\begin{table}[]
	\centering
	\caption{Statistics of the Six Benchmark Datasets}
	\label{datasets}
	\scalebox{0.8}{%
		\begin{tabular}{ccccc}
			\toprule
			Dataset & Subsets (Abbr.) & Samples & Feature (Size) & Classes \\ \midrule
			\multirow{3}{*}{Office31} & Amazon (A) & 2,817 & \multirow{3}{*}{Alexnet-FC$_7$ (4,096)} & \multirow{3}{*}{31} \\
			& DSLR (D) & 498 &  &  \\
			& Webcam (W) & 795 &  &  \\ \midrule
			\multirow{4}{*}{Office-Caltech10} & Amazon (A) & 958 & \multirow{4}{*}{\begin{tabular}[c]{@{}c@{}}SURF (800)\\ VGG-FC$_6$ (4,096)\\ VGG-FC$_7$ (4,096)\end{tabular}} & \multirow{4}{*}{10} \\
			& Caltech (C) & 1,123 &  &  \\
			& DSLR (D) & 157 &  &  \\
			& Webcam (W) & 295 &  &  \\ \midrule
			\multirow{5}{*}{PIE} & C05 & 3,332 & \multirow{5}{*}{Pixel (1024)} & \multirow{5}{*}{65} \\
			& C07 & 1,629 &  &  \\
			& C09 & 1,632 &  &  \\
			& C27 & 3,329 &  &  \\
			& C29 & 1,632 &  &  \\ \midrule
			\multirow{4}{*}{Office-Home} & Art (Ar) & 2,421 & \multirow{4}{*}{Resnet50 (2,048)} & \multirow{4}{*}{68} \\
			& Clipart (Cl) & 4,379 &  &  \\
			& Product (Pr) & 4,428 &  &  \\
			& Real-World (Re) & 4,357 &  &  \\ \midrule
			\multirow{2}{*}{MNIST-USPS} & MNIST (M) & 2,000 & \multirow{2}{*}{Pixel (256)} & \multirow{2}{*}{10} \\
			& USPS (U) & 1,800 &  &  \\ \midrule
			\multirow{2}{*}{ImageNet-VOC2007} & ImageNet (I) & 7,341 & \multirow{2}{*}{DeCAF$_6$ (4,096)} & \multirow{2}{*}{5} \\
			& VOC2007 (V) & 3,376 &  &  \\ \bottomrule
		\end{tabular}%
	}
\end{table}

\emph{PIE} \cite{PIE} involves 41,638 facial images of 68 people with different poses, illuminations, and expression changes. Following \cite{JDA}, we focus on five poses: C05 (left), C07 (upward), C09 (downward), C27 (frontal) and C29 (right). All images were converted to grayscale and cropped to the size 32 $\times$ 32. We adopt the pixel features\footnote{ \url{https://github.com/jindongwang/transferlearning/tree/master/data}\label{WJD}}.

\begin{table*}[]
	\centering
	\caption{Recognition Accuracies (\%) on Office31 Dataset}
	\label{office31}
	\scalebox{0.85}{%
		\begin{tabular}{ccccccccccccc}
			\toprule
			Task & 1-NN & SVM & JDA & DICD & PACET & MCS & DTLC & ARTL & MEDA & DGA-DA & DICE$_\mathrm{lp}$ & CDGS  \\ \midrule
			A$\rightarrow$D & 59.4 & 59.2 & 65.7 & 66.7 & 69.1 & 71.9 & 66.1 & 64.7 & 69.5 & 64.5 & 67.7 & \textbf{73.5}\\
			A$\rightarrow$W & 57.5 & 57.9 & 69.1 & 70.7 & 71.7 & 75.1 & 67.5 & 71.7 & 69.9 & 65.0 & 70.7 & \textbf{79.5}\\
			D$\rightarrow$A & 47.2 & 48.8 & 57.4 & 57.7 & \textbf{62.3} & 58.8 & 58.9 & 59.5 & 58.0 & 55.0 & 56.5 & 61.8\\
			D$\rightarrow$W & 96.1 & 95.2 & \textbf{98.0} & 97.0 & 97.4 & 96.7 & \textbf{98.0} & 96.0 & 94.0 & 97.2 & 97.2 & 97.2\\
			W$\rightarrow$A & 44.8 & 46.5 & 54.1 & 58.1 & 59.2 & 57.2 & 55.1 & 58.5 & 56.0 & 53.8 & 57.7 & \textbf{61.2}\\
			W$\rightarrow$D & 99.0 & 98.8 & 99.6 & 99.8 & \textbf{100.0} & 99.4 & 99.6 & 99.4 & 96.8 & 99.8 & \textbf{100.0} & \textbf{100.0} \\ \midrule
			Average & 67.3 & 67.7 & 74.0 & 75.0 & 76.6 & 76.5 & 74.2 & 74.9 & 74.0 & 72.5 & 75.0 & \textbf{78.9} \\ \bottomrule
		\end{tabular}%
	}
\end{table*}

\emph{MNIST-USPS} is made up of two handwritten digit image datasets: MNIST (M) and USPS (U). Following \cite{JDA}, we randomly choose 2,000 images in MNIST and 1,800 images in USPS and utilize the pixel features\textsuperscript{\ref{WJD}}.

\emph{ImageNet-VOC2007} consists of two large image recognition datasets, ImageNet (I) and VOC2007 (V). Following \cite{MEDA}, we extract all images from five common classes of the two datasets, {\it i.e.}, bird, cat, chair, dog and person. The DeCAF$_6$  feature\textsuperscript{\ref{WJD}} is employed.

\emph{Office-Home} \cite{Office-Home} includes 15,585 object images in 65 categories from four domains: Art (artistic depictions of objects, Ar), Clipart (clipart images, Cl), Product (object images without background, Pr) and Real-World (images captured by a regular camera, Re). We employ the Resnet50 features extracted by a Resnet50 model \cite{Resnet50} pretrained on ImageNet.

For simplicity, in our experiments, each cross-domain task is denoted by S$\rightarrow$ T, where S represents the source domain and T is the target domain.
\subsection{Experimental Setup}
\subsubsection{Comparison Methods}
For UDA, we compare the performance of our CDGS with massive methods, which can be classified into two categories: \emph{shallow methods}: 1-NN, SVM\footnote{\url{https://www.csie.ntu.edu.tw/~cjlin/liblinear/}}, JDA \cite{JDA}, DICD \cite{DICD}, PACET \cite{PACET}, MCS \cite{MCS}, DTLC \cite{DTLC}, ARTL \cite{ARTL}, MEDA \cite{MEDA}, DGA-DA \cite{DGA-DA} and DICE$_\mathrm{lp}$ \cite{DICE}, \emph{deep methods}: the method of \cite{Wang2020}, DRCN \cite{DRCN}, DSAN \cite{DSAN}, the method of \cite{Liang2020}, and GSP \cite{GSP}. For SDA, the competitors include MMDT \cite{MMDT}, CDLS \cite{CDLS}, ILS \cite{ILS}, TFMKL-S \cite{TFMKL-S} and OBTL \cite{OBTL}.

\subsubsection{Training Protocol}
We exploit all source data for training, known as full protocol, on all datasets in Table \ref{datasets}. Besides, regarding the Office-Caltech10 dataset, two kinds of sampling protocols are also adopted, where only few labeled source samples per category are employed for training. For the first sampling protocol, similar to \cite{DICE}, we use the SURF features and 20 instances per class are randomly selected for domain A while 8 instances per class for other domains as sources. For the second sampling protocol, following \cite{MCS}, VGG-FC$_6$ features are utilized and 8 samples per category are selected for domain D while 20 samples per category for the others.

\subsubsection{Parameter Setting}
In UDA and SDA, sufficient labeled target samples are unavailable, thus we cannot perform a standard cross-validation procedure to decide the optimal parameters. Following \cite{DICD}, we report the best results by grid-searching the hyper-parameter space. For all competitors, we run the public codes provided by the authors using the default parameters or following the given procedure to tune parameters. For all approaches requiring a subspace dimension, the optimal value is searched in $d \in \{1C, 2C, 3C, 4C, 5C, 6C\}$, where $C$ is the number of classes for the corresponding dataset. The regulation parameter for projection matrix is searched in  $\gamma \in \{0.005,0.01,0.05,0.1,0.5,1.0,5.0,10.0\}$. For the other parameters in our CDGS, we fix $\alpha = 1.0$, $k = 20$, $\delta = 0.8$, $T = 10$ and set $\beta = 0.5$ for Office-Home and Office-Caltech10 datasets, $\beta = 0.01$ for PIE dataset and $\beta = 0.1$ for other datasets. We also provide the optimal parameters for UDA setting: Office31 ($d = 124$, $\gamma = 0.01$), Office-Caltech10 ($d = 30$, $\gamma = 0.5$ for SURF, $d = 30$, $\gamma = 0.1$ for SURF split, $d = 40$, $\gamma = 0.1$ for VGG-FC$_{6,7}$ split), MNIST-USPS ($d = 40$, $\gamma = 0.5$), ImageNet-VOC2007 ($d = 30$, $\gamma = 0.01$), PIE ($d = 340$, $\gamma = 0.005$) and Office-Home ($d = 130$, $\gamma = 0.005$).
\begin{table*}[]
	\centering
	\caption{Recognition Accuracies (\%) on Office-Caltech10 Dataset with SURF Features}
	\label{Office_surf}
	\scalebox{0.85}{%
		\begin{tabular}{ccccccccccccc}
			\toprule
			Task & 1-NN & SVM & JDA & DICD & PACET & MCS & DTLC  & ARTL & MEDA & DGA-DA & DICE$_\mathrm{lp}$ & CDGS \\ \midrule
			A$\rightarrow$C & 26.0 & 35.6 & 39.4 & 42.4 & 42.7  & 40.8 & \textbf{46.6}  & 41.3 & 43.9 & 41.3 & 44.1 & 42.7 \\
			A$\rightarrow$D & 25.5 & 36.3 & 39.5 & 38.9 & 50.3  & 45.2 & 45.4  & 38.9 & 45.9 & 38.2 & 49.0 & \textbf{51.0} \\
			A$\rightarrow$W & 29.8 & 31.9 & 38.0 & 45.1 & \textbf{53.2}  & 50.8 & 48.1  & 39.0 & \textbf{53.2} & 38.3 & 52.9 & 52.5\\
			C$\rightarrow$A & 23.7 & 42.9 & 44.8 & 47.3 & 52.2  & \textbf{58.8} & 50.3  & 54.9 & 56.5 & 52.1 & 53.7 & 56.8 \\
			C$\rightarrow$D & 25.5 & 33.8 & 45.2 & 49.7 & 52.2  & 45.2 & 52.4  & 44.6 & 50.3 & 45.9 & 51.6 & \textbf{59.2} \\
			C$\rightarrow$W & 25.8 & 34.6 & 41.7 & 46.4 & 51.5 & 51.9 & 54.4  & 50.5 & 53.9 & 47.1 & 53.9 &\textbf{55.9} \\
			D$\rightarrow$A & 28.5 & 34.3 & 33.1 & 34.5 & 40.8 & 37.1 & 36.2  & 38.1 & 41.2 & 33.6 & 41.2 & \textbf{45.2} \\
			D$\rightarrow$C & 26.3 & 32.1 & 31.5 & 34.6 & 34.5  & 31.3 & 32.1  & 31.0 & 34.9 & 33.7 & 34.5 & \textbf{39.4} \\
			D$\rightarrow$W & 63.4 & 78.0 & 89.5 & 91.2 & 91.5  & 86.1 & 92.9  & 83.4 & 87.5 & \textbf{93.2} & 84.1 & 92.5 \\
			W$\rightarrow$A & 23.0 & 37.5 & 32.8 & 34.1 & 40.8  & 37.8 & 33.5  & 40.1 & 42.7 & 41.8 & 33.1 & \textbf{47.4} \\
			W$\rightarrow$C & 19.9 & 33.9 & 31.2 & 33.6 & \textbf{39.0} & 29.8 & 33.8  & 34.8 & 34.0 & 33.3 & 37.8 & 38.2 \\
			W$\rightarrow$D & 59.2 & 80.9 & 89.2 & 89.8 & 92.4  & 83.4 & 87.3  & 78.3 & 88.5 & 89.8 & 87.3 & \textbf{94.3} \\ \midrule
			Average & 31.4 & 42.6 & 46.3 & 49.0 & 53.6  & 50.1 & 51.1  & 47.9 & 52.7 & 49.0 & 51.9 & \textbf{56.3} \\ \bottomrule
		\end{tabular}%
	}
\end{table*}

\begin{table*}[]
	\centering
	\caption{Recognition Accuracies (\%) on Office-Caltech10 Dataset with SURF and VGG-FC$_{6,7}$ Features  under Different Splitting Protocols}
	\label{Office_surf_split}
	\scalebox{0.8}{%
		\begin{tabular}{c|ccccc|ccccc|ccccc}
			\toprule
			Feature & \multicolumn{5}{c|}{SURF} & \multicolumn{5}{c|}{VGG-FC$_6$} & \multicolumn{5}{c}{VGG-FC$_7$} \\ \midrule
			Task & MCS & ARTL & MEDA & DICE$_\mathrm{lp}$ & CDGS & MCS & ARTL & MEDA & DICE$_\mathrm{lp}$ & CDGS & MCS & ARTL & MEDA & DICE$_\mathrm{lp}$ & CDGS \\ \midrule
			A$\rightarrow$C & 40.4 & 36.4 & 38.1 & 39.6 & \textbf{40.8} & \textbf{87.1} & 84.6 & 85.2 & 83.9 & 85.1 & \textbf{86.3} & 84.1 & 84.4 & 83.6 & 84.8 \\
			A$\rightarrow$D & 43.7 & 38.2 & 39.0 & 39.7 & \textbf{44.3} & 74.8 & 75.0 & 71.7 & 66.4 & \textbf{82.6} & 72.8 & 74.8 & 70.8 & 64.9 & \textbf{81.0} \\
			A$\rightarrow$W & 48.3 & 38.3 & 45.8 & 42.9 & \textbf{49.8} & 84.8 & 90.1 & 88.5 & 77.2 & \textbf{97.9} & 86.6 & 87.8 & 88.2 & 79.4 & \textbf{94.5} \\
			C$\rightarrow$A & 43.2 & 41.5 & 44.3 & \textbf{44.8} & 44.5 & \textbf{92.3} & 89.0 & 90.8 & 91.6 & 88.8 & \textbf{92.8} & 91.1 & 91.6 & 91.6 & 88.9 \\
			C$\rightarrow$D & 45.3 & 39.0 & 39.9 & 39.6 & \textbf{46.7} & 77.3 & 79.9 & 78.1 & 68.2 & \textbf{84.5} & 73.0 & 79.0 & 75.7 & 65.4 & \textbf{83.5} \\
			C$\rightarrow$W & 43.8 & 35.8 & 40.3 & 40.5 & \textbf{44.7} & 87.1 & 89.9 & 90.3 & 83.8 & \textbf{93.9} & 89.3 & 90.6 & 90.2 & 84.2 & \textbf{93.0} \\
			D$\rightarrow$A & 37.7 & 37.0 & 40.3 & 40.9 & \textbf{45.0} & 84.7 & \textbf{90.4} & 86.3 & 85.0 & 84.8 & 84.6 & \textbf{88.9} & 83.7 & 83.3 & 85.7 \\
			D$\rightarrow$C & 30.8 & 32.1 & 33.5 & 33.8 & \textbf{35.6} & 76.0 & 75.5 & \textbf{81.2} & 77.9 & 71.5 & \textbf{76.5} & 74.9 & 73.8 & 76.0 & 70.8 \\
			D$\rightarrow$W & 78.1 & 79.7 & 82.5 & 81.7 & \textbf{86.2} & 95.9 & 95.7 & 96.1 & 95.7 & \textbf{96.9} & 95.5 & 94.3 & 95.8 & 94.8 & \textbf{96.9} \\
			W$\rightarrow$A & 36.3 & 37.6 & 40.9 & 38.7 & \textbf{44.0} & 88.9 & 92.0 & 90.6 & 89.8 & \textbf{92.2} & 90.4 & \textbf{92.8} & 90.2 & 89.4 & 92.5 \\
			W$\rightarrow$C & 32.6 & 32.3 & 33.1 & 34.5 & \textbf{35.4} & \textbf{87.4} & 85.6 & 85.2 & 81.0 & 85.6 & 85.6 & \textbf{85.8} & 84.6 & 82.3 & 85.0 \\
			W$\rightarrow$D & 73.3 & 68.5 & 74.6 & 76.1 & \textbf{78.8} & 92.9 & 92.8 & \textbf{97.2} & 93.9 & 96.7 & 88.9 & 91.1 & 93.4 & 90.8 & \textbf{93.5} \\ \midrule
			Averaged & 46.1 & 42.9 & 46.0 & 46.1 & \textbf{49.7} & 85.8 & 86.7 & 86.8 & 82.9 & \textbf{88.2} & 85.2 & 86.3 & 85.2 & 82.1 & \textbf{87.5} \\ \bottomrule
		\end{tabular}%
	}
\end{table*}

\subsubsection{Evaluation Metric}
Following many previous works \cite{JDA,DICD,DICE}, we adopt the classification accuracy of target data as the evaluation metric, which is computed as:
\begin{equation}
\label{evaluation}
\mathrm{Accuracy} = \frac{|\mathbf{x}:\mathbf{x} \in \mathbf{X}_t \cap \tilde{y} = y|}{|\mathbf{x}:\mathbf{x} \in \mathbf{X}_t|}
\end{equation}
where $\mathbf{x}$ is a target sample, $y$ is the truth label of $\mathbf{x}$, and $\tilde{y}$ is the corresponding pseudo-label.

\subsection{Unsupervised Domain Adaptation}
\subsubsection{The Experimental Results on Unsupervised Domain Adaptation}
{\it a) Results on Office31 Dataset}. The classification accuracies of all methods on this dataset are listed in Table \ref{office31}, where the highest accuracy for each task is boldfaced. The results of DGA-DA are copied from \cite{DICE}. It is observed that CDGS performs much better than all competitors. Specifically, CDGS achieves 78.9$\%$ average accuracy, which leads the second best method PACET by 2.3$\%$. DICE$_\mathrm{lp}$ and DGA-DA both explore the geometric structure underlying data manifold to assign target pseudo-labels by cross-domain label propagation. However, CDGS further integrates domain-invariant feature learning, affinity matrix constructing and target labels inferring into one framework. Therefore, CDGS could make the three parts interact with each other to yield a superior performance. Besides, CDGS employs a self-learning strategy to construct a discriminative graph to capture the inherent similarity of samples as well as explore the label information of source and target data. In such case, the discriminative graph can transfer source knowledge to target domain more effectively.

{\it b) Results on Office-Caltech10 Dataset.} The results on  Office-Caltech10 dataset with SURF features under the full protocol are shown in table \ref{Office_surf}. In terms of the average accuracy, CDGS owns a large advantage, which improves 2.7$\%$ over the best competitor PACET. CDGS  works the best for 7 out of 12 tasks while PACET only wins two tasks, which verifies the significant effectiveness of CDGS. Compared with these methods which employ cross-domain label propagation to  infer target labels, {\it i.e.}, ARTL, MEDA, DGA-DA and DICE$_\mathrm{lp}$, the improvement of CDGS is 3.6$\%$, which illustrates the superiority of our CDGS over the counterparts.

Then, we also compare our CDGS with several competitors under different splitting protocols with different features. The results over 20 random splits are illustrated in table \ref{Office_surf_split}. For SURF features, CDGS performs much better than other methods in terms of the average accuracy. CDGS achieves 49.7$\%$ average performance, which owns 3.6$\%$ improvement compared with the best competitors, MCS and DICE$_\mathrm{lp}$. Notably, CDGS performs the best on all tasks except for C$\rightarrow$A. For VGG-FC$_{6,7}$ features, CDGS  outperforms all comparison methods again. Carefully comparing the results of SURF and VGG-FC$_{6,7}$ features, we can find that CDGS can consistently achieve good performance regardless of the features, which illustrates that CDGS holds better generalization capacity.

{\it c) Results on MNIST-USPS and ImageNet-VOC2007 Datasets.} To verify the effectiveness of CDGS on digit images, we further conduct experiments on MNIST-USPS dataset. The comparison results are listed in Table \ref{MI}. CDGS achieves the highest average accuracy compared with all competitors. We can observe that CDGS is much superior to feature adaptation approaches, {\it e.g.}, DGA-DA and DICE$_\mathrm{lp}$, and owns 5.7$\%$ improvement in terms of the average accuracy, which demonstrates the superiority of our proposal. The classification results of all methods on ImageNet-VOC2007 dataset are also provided in Table \ref{MI}. CDGS performs much better than other methods. Moreover, compared with the related methods, {\it i.e.}, ARTL, MEDA and DICE$_\mathrm{lp}$, CDGS shows large improvement up to 8.3$\%$, which confirms the advancement of our CDGS. 

\begin{table*}[]
	\centering
	\caption{Recognition Accuracies (\%) on MNIST-USPS and  ImageNet-VOC2007 Datasets. ``$-$" Indicates That the Results Are Unavailable}
	\label{MI}
	\scalebox{0.85}{%
		\begin{tabular}{ccccccccccccccc}
			\toprule
			Task & 1-NN & SVM & JDA & DICD & PACET  & MCS & DTLC & ARTL & MEDA & DGA-DA & DICE$_\mathrm{lp}$ & CDGS  \\ \midrule
			M$\rightarrow$U & 65.9 & 50.0 & 67.3 & 77.8 & 77.0  & 53.4 & 70.7 & 88.5 & \textbf{89.5} & 82.3 & 78.3 & 87.1 \\
			U$\rightarrow$M & 44.7 & 29.9 & 59.7 & 65.2 & 64.5 & 51.3 & 56.9 & 61.8 & 72.1 & 70.8 & 65.2 & \textbf{77.3} \\ \midrule
			Average & 55.3 & 40.0 & 63.5 & 71.5 & 70.7  & 52.4 & 63.8 & 75.1 & 80.8 & 76.5 & 71.8 & \textbf{82.2}  \\ \toprule
			I$\rightarrow$V & 65.4 & 69.9 & 63.9 & 64.5 & 63.9  & 60.6 & 64.8 & 65.4 & 67.3 & $-$ & 65.8 & \textbf{72.3}  \\
			V$\rightarrow$I & 73.7 & 78.7 & 72.4 & 78.2 & 72.8  & 84.2 & 85.8 & 77.8 & 74.7 & $-$ & 76.0 & \textbf{87.5} \\ \midrule
			Average & 69.5 & 74.3 & 68.1 & 71.3 & 68.4  & 72.4 & 75.3 & 71.6 & 71.0 & $-$ & 70.9 & \textbf{79.9} \\ \bottomrule
		\end{tabular}%
	}
\end{table*}

\begin{table*}[]
	\centering
	\caption{Recognition Accuracies (\%) on PIE Dataset}
	\label{PIE}
	\scalebox{0.85}{%
		\begin{tabular}{cccccccccccccc}
			\toprule
			Task & 1-NN & SVM & JDA & DICD & PACET  & MCS & DTLC & ARTL & MEDA & DGA-DA & DICE$_\mathrm{lp}$ & CDGS\\ \midrule
			C05$\rightarrow$C07 & 26.1 & 30.9 & 58.8 & 73.0 & 82.2  & 68.9 & \textbf{85.1} & 59.7 & 64.2 & 65.3 & 83.9 & 84.8 \\
			C05$\rightarrow$C09 & 26.6 & 33.9 & 54.2 & 72.0 & 80.8  & 62.9 & \textbf{82.7} & 57.8 & 59.4 & 62.8 & 77.5 & 81.4 \\
			C05$\rightarrow$C27 & 30.7 & 41.4 & 84.5 & 92.2 & 94.9  & 87.9 & 97.1 & 86.3 & 84.2 & 83.5 & 95.9 & \textbf{98.5} \\
			C05$\rightarrow$C29 & 16.7 & 23.8 & 49.8 & 66.9 & 64.5  & 53.6 & \textbf{77.2} & 47.5 & 46.5 & 56.1 & 66.0 & 72.9 \\
			C07$\rightarrow$C05 & 24.5 & 31.8 & 57.6 & 69.9 & \textbf{82.9}  & 62.3 & 82.8 & 78.3 & 77.5 & 63.7 & 81.4 & 80.0 \\
			C07$\rightarrow$C09 & 46.6 & 41.0 & 62.9 & 65.9 & 73.5  & 52.1 & 83.9 & 69.0 & 71.4 & 61.3 & 74.1 & \textbf{86.5} \\
			C07$\rightarrow$C27 & 54.1 & 62.2 & 75.8 & 85.3 & 90.1  & 80.8 & 92.1 & 90.0 & 89.2 & 82.4 & 88.4 & \textbf{93.6} \\
			C07$\rightarrow$C29 & 26.5 & 28.8 & 39.9 & 48.7 & 72.4  & 58.8 & 79.7 & 54.2 & 58.6 & 46.6 & 68.0 & \textbf{81.4} \\
			C09$\rightarrow$C05 & 21.4 & 32.3 & 51.0 & 69.4 & 79.7  & 58.4 & 80.0 & 72.3 & 73.2 & 56.7 & 78.0 & \textbf{85.3} \\
			C09$\rightarrow$C07 & 41.0 & 39.7 & 58.0 & 65.4 & 79.3  & 52.5 & \textbf{84.4} & 66.5 & 68.8 & 61.3 & 75.9 & 82.6 \\
			C09$\rightarrow$C27 & 46.5 & 61.9 & 68.5 & 83.4 & 84.6  & 82.8 & 94.3 & 85.5 & 86.9 & 77.8 & 85.2 & \textbf{95.6} \\
			C09$\rightarrow$C29 & 26.2 & 37.7 & 40.0 & 61.4 & 70.2  & 59.1 & 79.9 & 60.0 & 65.6 & 44.2 & 71.3 & \textbf{81.0} \\
			C27$\rightarrow$C05 & 33.0 & 57.7 & 80.6 & 93.1 & 94.0  & 87.7 & 96.7 & 88.7 & 89.6 & 81.8 & 93.3 & \textbf{98.8} \\
			C27$\rightarrow$C07 & 62.7 & 69.2 & 82.6 & 90.1  & 93.5 & 87.2 & 94.8 & 86.7 & 88.6 & 85.3 & 95.0 & \textbf{95.2} \\
			C27$\rightarrow$C09 & 73.2 & 69.7 & 87.3 & 89.0 & 91.3  & 83.6 & \textbf{95.4} & 87.6 & 88.8 & 91.0 & 92.3 & 93.9 \\
			C27$\rightarrow$C29 & 37.2 & 48.7 & 54.7 & 75.6 & 77.0  & 79.2 & 84.4 & 71.0 & 78.3 & 53.8 & 80.5 & \textbf{88.6} \\
			C29$\rightarrow$C05 & 18.5 & 29.4 & 46.5 & 62.9 & 76.2  & 48.7 & 75.4 & 66.1 & 65.2 & 57.4 & 74.2 & \textbf{84.4} \\
			C29$\rightarrow$C07 & 24.2 & 33.1 & 42.1 & 57.0 & 69.2 & 58.4 & \textbf{77.8} & 57.3 & 58.1 & 53.8 & 69.2 & 75.7 \\
			C29$\rightarrow$C09 & 28.3 & 40.6 & 53.3 & 65.9 & 79.2  & 63.4 & 82.4 & 62.9 & 68.1 & 55.3 & 74.6 & \textbf{83.5} \\
			C29$\rightarrow$C27 & 31.2 & 51.5 & 57.0 & 74.8 & 85.3 & 76.2 & \textbf{89.7} & 76.2 & 78.0 & 61.8 & 83.5 & 89.5\\ \midrule
			Average & 34.8 & 43.3 & 60.3 & 73.1 & 81.0  & 68.2 & 85.8 & 71.2 & 73.0 & 65.1 & 80.4 & \textbf{86.7} \\ \bottomrule
		\end{tabular}%
	}
\end{table*}

\begin{table*}[]
	\centering
	\caption{Recognition Accuracies (\%) on Office-Home Dataset. Deep Learning Methods Are Below CDGS}
	\label{Office-Home}
	\scalebox{0.85}{%
		\begin{tabular}{cccccccccccccc}
			\toprule
			Method & Ar$\rightarrow$Cl & Ar$\rightarrow$Pr & Ar$\rightarrow$Re & Cl$\rightarrow$Ar & Cl$\rightarrow$Pr & Cl$\rightarrow$Re & Pr$\rightarrow$Ar & Pr$\rightarrow$Cl & Pr$\rightarrow$Re & Re$\rightarrow$Ar & Re$\rightarrow$Cl & Re$\rightarrow$Pr & Average \\ \midrule
			1-NN & 38.0 & 54.4 & 61.5 & 40.7 & 52.6 & 52.6 & 47.1 & 41.0 & 66.7 & 57.1 & 45.1 & 73.2 & 52.5 \\
			SVM & 47.1 & 66.2 & 73.3 & 50.8 & 62.1 & 63.9 & 54.0 & 44.4 & 73.6 & 62.5 & 47.9 & 77.3 & 60.3 \\
			JDA & 45.3 & 62.5 & 65.7 & 51.9 & 62.1 & 62.2 & 55.0 & 47.2 & 71.8 & 60.6 & 50.3 & 73.8 & 59.0 \\
			DICD & 46.4 & 63.3 & 68.4 & 53.7 & 62.7 & 64.4 & 56.0 & 45.6 & 72.0 & 63.2 & 50.2 & 76.7 & 60.2 \\
			PACET & 52.1 & 71.5 & 76.3 & 62.2 & 75.4 & 75.5 & 60.8 & 50.5 & 79.3 & 67.8 & 56.6 & 81.5 & 67.4 \\
			MCS & 54.2 & 76.4 & 78.9 & 63.7 & 74.5 & 78.3 & 55.9 & 53.2 & 79.9 & 68.1 & 55.9 & 80.2 & 69.1 \\
			DTLC & 51.9 & 74.1 & 75.1 & 61.4 & 70.7 & 73.7 & 63.0 & 51.1 & 76.1 & 66.7 & 54.9 & 79.0 & 66.5 \\
			ARTL & 52.9 & 73.8 & 76.7 & 63.0 & \textbf{78.5} & 77.1 & 63.6 & 52.4 & 78.7 & 70.1 & 55.2 & 82.6 & 68.7 \\
			MEDA & 52.9 & 75.7 & 77.4 & 60.3 & 77.6 & 77.8 & 62.5 & 52.8 & 79.3 & 68.4 & 54.7 & 82.4 & 68.5 \\
			DICE$_\mathrm{lp}$ & 48.4 & 70.8 & 72.7 & 52.9 & 65.2 & 65.6 & 59.3 & 49.0 & 76.4 & 65.1 & 52.6 & 79.0 & 63.1 \\
			CDGS & 55.6 & \textbf{77.0} & \textbf{80.1} & \textbf{67.1} & 77.2 & \textbf{78.4} & \textbf{66.5} & 54.4 & \textbf{81.0} & 70.5 & 57.9 & 83.3 & \textbf{70.8} \\ \midrule
			Wang \emph{et al}. \cite{Wang2020} & 55.5 & 73.5 & 78.7 & 60.7 & 74.1 & 73.1 & 59.5 & 55.0 & 80.4 & 72.4 & 60.3 & \textbf{84.3} & 68.9 \\
			DRCN & 50.6 & 72.4 & 76.8 & 61.9 & 69.5 & 71.3 & 60.4 & 48.6 & 76.8 & 72.9 & 56.1 & 81.4 & 66.6 \\
			DSAN & 54.4 & 70.8 & 75.4 & 60.4 & 67.8 & 68.0 & 62.6 & \textbf{55.9} & 78.5 & \textbf{73.8} & \textbf{60.6} & 83.1 & 67.6 \\
			Liang \emph{et al.} \cite{Liang2020} & 54.1 & 74.2 & 77.7 & 62.9 & 73.6 & 74.6 & 63.4 & 54.9 & 80.4 & 73.1 & 58.2 & 83.6 & 69.2 \\
			GSP & \textbf{56.8} & \textbf{75.5} & 78.9 & 61.3 & 69.4 & 74.9 & 61.3 & 52.6 & 79.9 & 73.3 & 54.2 & 83.2 & 68.4 \\ \bottomrule
		\end{tabular}%
	}
\end{table*}

{\it d) Results on PIE Dataset.} Table \ref{PIE} summarizes the classification performance of CDGS and other methods on PIE dataset. We can observe that CDGS performs better than all competitors in terms of the average performance. Specifically, CDGS achieves the highest average classification accuracy, which owns 0.9$\%$ improvement against the best competitor DTLC. Besides, CDGS wins 12 out of 20 tasks while  DTLC only performs the best on 7 tasks. It is worthy to note that compared with ARTL, MEDA, DGA-DA and DICE$_\mathrm{lp}$, CDGS achieves 6.3$\%$ improvement, which indicates that our CDGS is more conductive for cross-domain face recognition tasks.

{\it e) Results on Office-Home Dataset.} For this large-scale dataset, we use the Resnet50 model pretrained on ImageNet to extract features. The classification results are shown in Table \ref{Office-Home}. Here, we also report the results of five recent deep domain adaptation methods, which take the Resnet50 model as the backbone. It is clearly observed that our CDGS outperforms all traditional and deep comparison methods in average accuracy.
Specifically, CDGS leads the best traditional competitor MCS by 1.7$\%$. In addition, CDGS is the best method on 5 out of 12 tasks while MCS only wins one task, which verifies the significant effectiveness of our proposal against the traditional competitors. Compared with the best deep competitor,  CDGS achieves 1.6$\%$ improvement, which validates the superiority of our proposal when equipped with off-the-self deep features.

For a complete understanding, we summarize the average accuracy of several competitors and our CDGS on all benchmark datasets under the full protocol in Table \ref{averaged accuracy}. We  discover that CDGS obtains the highest average accuracy, leading the best competitor MEDA by 5.8$\%$, which validates that our CDGS is capable of addressing various DA tasks effectively. 
\begin{table}[h]
	\centering
	\caption{Average Accuracies of Our CDGS and Several Competitors on All Six Datasets under the Full Protocol}
	\label{averaged accuracy}
	\scalebox{0.85}{%
		\begin{tabular}{ccccccc}
			\midrule
			Method & MCS & DTLC & ARTL & DICE$_\mathrm{lp}$ & MEDA & CDGS \\ \midrule
			Average accuracy & 64.7 & 69.5 & 68.2 & 68.9 & 70.0 & \textbf{75.8}\\ \midrule
		\end{tabular}%
	}
\end{table}

\subsubsection{Ablation Study} 
To understand our method more deeply, we propose three variants of CDGS:
\emph{a}) CDGS$^\mathrm{sp}$, Separates domain-invariant feature learning, affinity matrix constructing and target labels inferring into three independent stages and constructs the affinity matrix with Predefined  similarity metric, {\it i.e.}, the gaussian kernel similarity with kernel width 1.0; \emph{b}) CDGS$^\mathrm{dg}$, integrates Domain-invariant feature learning and Graph self-learning into one framework, {\it i.e.}, jointing Eq. (\ref{MMD}), Eq. (\ref{asm}) and Eq. (\ref{reg}); \emph{c}) CDGS$^\mathrm{ds}$, jointly performs Domain-invariant feature learning and graph self-learning with Source domain discriminative structure preserving, {\it i.e.}, unifying Eq. (\ref{MMD}), Eq. (\ref{asm1}) and Eq. (\ref{reg}). It is worthy noting that compared with CDGS$^\mathrm{ds}$, our CDGS further considers the label smoothness constraint during the  discriminative graph self-learning. In Table \ref{ablation study}, we list the average classification accuracy of CDGS and the three variants on all datasets under the full protocol. Based on this table, more detailed analysis about our CDGS is presented as follows.

{\it a) Effectiveness of Graph Self-learning.} As we can see, CDGS$^\mathrm{dg}$ is superior to CDGS$^\mathrm{sp}$ on all datasets except for PIE, which verifies the effectiveness of graph self-learning. Particularly, compared with CDGS$^\mathrm{sp}$, CDGS$^\mathrm{dg}$ achieves 5.9$\%$ improvement on ImageNet-VOC2007 dataset and 3.3$\%$ improvement on Office-Home dataset respectively, which confirms the superiority of graph self-learning.  By integrating the domain-invariant feature learning and graph self-learning into one framework, we can capture the inherent similarity  connections among source and target samples more effectively, and thus improve the classification performance of cross-domain label propagation. 

{\it b) Effectiveness of Graph Self-learning with Source Discriminative Structure Preserving.} We can see that CDGS$^\mathrm{ds}$ performs much better than CDGS$^\mathrm{dg}$ in terms of average accuracy, which achieves a large improvement of 5.1$\%$. Notably, on  datasets MNIST-USPS and PIE, CDGS$^\mathrm{ds}$ even achieves more than 12.9$\%$ advancement.  The above results demonstrate that  preserving the source discriminative structure in graph self-learning process is of vital importance to improve the quality of affinity matrix, such that the knowledge from source domain can be transferred to target domain more effectively. 

{\it c) Effectiveness of Label Smoothness Constraint for Discriminative Graph Self-learning.} It is observed that our CDGS  outperforms CDGS$^\mathrm{ds}$ on 5 out of all 6 datasets and achieves superior performance in terms of average accuracy. This phenomenon indicates that
the introduction of weakly supervised information contained in target pseudo-labels can promote to yield a discriminative graph with higher quality, and thus the source knowledge can be propagated to target domain more adequately.

\begin{figure*}[]
	\setlength{\abovecaptionskip}{0pt}
	\setlength{\belowcaptionskip}{0pt}
	\renewcommand{\figurename}{Figure}
	\centering
	\includegraphics[width=1.0\textwidth]{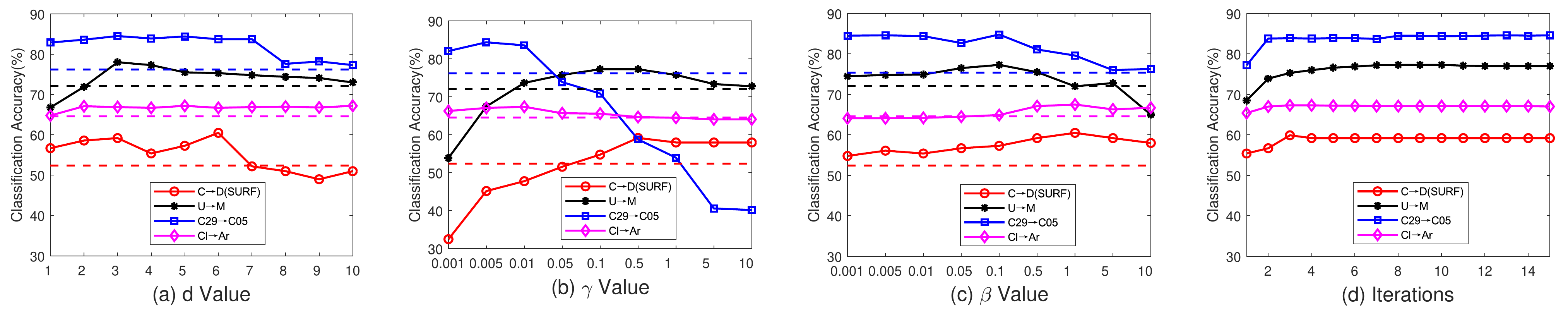}
	\caption{Parameter Sensitivity Analysis of CDGS with respect to $d$, $\gamma$, $\beta$ and $T$.}
	\label{parameter_analysis}
\end{figure*}

\begin{table}[]
	\centering
	\caption{The Average Classification Accuracies of CDGS and the Three Variants on All Datasets under the Full Protocol}
	\label{ablation study}
	\scalebox{0.85}{%
		\begin{tabular}{p{2.4cm}<{\centering}p{1.4cm}<{\centering}p{1.4cm}<{\centering}p{1.4cm}<{\centering}p{1.4cm}<{\centering}}
			\toprule
			Dataset &  CDGS$^\mathrm{sp}$ & CDGS$^\mathrm{dg}$ & CDGS$^\mathrm{ds}$  & CDGS \\ \midrule
			Office31  & 74.5  & 75.6 & 78.5  & \textbf{78.9} \\
			Office-Caltech10  & 53.3  & 54.5 & 55.2  & \textbf{56.3} \\
			MNIST-USPS  & 66.0  & 66.7 & 80.4  & \textbf{82.2} \\
			ImageNet-VOC2007  & 71.7  & 77.6 & 76.6  & \textbf{79.9} \\
			PIE  & 75.2  & 73.9 & \textbf{86.8}   & 86.7 \\
			Office-Home  & 65.3  & 68.6 & 69.8  & \textbf{70.8} \\ \midrule
			Average  & 67.7  & 69.5 & 74.6  & \textbf{75.8} \\  \bottomrule
		\end{tabular}%
	}
\end{table}

\begin{table*}[]
	\centering
	\caption{Recognition Accuracies (\%) on Office-Caltech10 Dataset with SURF Features for SDA.}
	\label{office_surf_semi}
	\scalebox{0.85}{%
		\begin{tabular}{cccccccccccccc}
			\toprule
			Method & A$\rightarrow$C & A$\rightarrow$D & A$\rightarrow$W & C$\rightarrow$A & C$\rightarrow$D & C$\rightarrow$W & D$\rightarrow$A & D$\rightarrow$C & D$\rightarrow$W & W$\rightarrow$A & W$\rightarrow$C & W$\rightarrow$D & Averaged \\ \midrule
			MMDT & 36.4 & 56.7 & 64.6 & 49.4 & 56.5 & 63.8 & 46.9 & 34.1 & 74.1 & 47.7 & 32.2 & 67.0 & 52.5 \\
			CDLS & 35.3 & 60.4 & 68.7 & 50.9 & 59.8 & 66.3 & 50.7 & 34.9 & 68.5 & 51.8 & 33.5 & 60.7 & 53.5 \\
			ILS & 43.6 & 49.8 & 68.7 & 55.1 & 56.2 & 62.9 & 55.0 & \textbf{41.0} & 80.1 & 54.3 & 38.6 & 70.8 & 55.6 \\
			TFMKL-S & 43.8 & \textbf{62.0} & 70.9 & 54.2 & 60.1 & 68.1 & 53.1 & 38.9 & 79.1 & 54.4 & 36.2 & 69.1 & 57.5 \\
			OBTL & 41.5 & 60.2 & 72.4 & 54.8 & 56.2 & 71.1 & 54.4 & 40.3 & 83.2 & \textbf{55.0} & 37.4 & 75.0 & 58.9 \\
			CDGS & \textbf{44.7} & 60.9 & \textbf{73.5} & \textbf{57.9} & \textbf{63.9} & \textbf{75.0} & \textbf{57.3} & 40.8 & \textbf{87.1} & 54.1 & \textbf{39.4} & \textbf{79.3} & \textbf{61.1} \\ \bottomrule
		\end{tabular}%
	}
\end{table*}

\begin{table}[]
	\centering
	\caption{Recognition Accuracies (\%) on MNIST-USPS Dataset for SDA.}
	\label{mnist_semi}
	\scalebox{0.9}{%
		\begin{tabular}{cccccc}
			\toprule
			Task    & MMDT & CDLS & ILS & TFMKL-S & CDGS\\ \midrule
			M$\rightarrow$U     &61.7  &79.0  &45.2  &69.5  &\textbf{88.0}   \\
			U$\rightarrow$M     &47.8  &68.6  &50.4  &53.5  &\textbf{79.1} \\ \midrule
			Averaged   &54.8    &73.8  &47.8  &61.5  &\textbf{83.5}   \\ \bottomrule
		\end{tabular}%
	}
\end{table}

\subsubsection{Parameter Sensitivity and Convergence Analysis}
Three tunable parameters are involved in our CDGS: $d$, $\gamma$, $\beta$. We have conducted extensive parameter sensitivity analysis on object, digit and face datasets by varying one parameter once in a wide range and fixing the other parameters to the optimal values. We display the results of task C$\rightarrow$ D (SURF), U$\rightarrow$M, C29$\rightarrow$C05 and Cl$\rightarrow$Pr in Fig. \ref{parameter_analysis} (a) $\sim$ (c). To verify the effectiveness of our CDGS,  the results of the best competitor for each task are also provided as the dash lines.

First, we run CDGS as $d$ varies in $d \in [1C,2C,...,10C]$, where $C$ is the number of classes for the corresponding task. From Fig. \ref{parameter_analysis} (a), we can observe that our CDGS is robust to different values of $d$. We empirically find that $d \in [2C,7C]$ is an optimal choice. Then, we investigate the sensitivity of $\gamma$ by varying it from 0.001 to 10.0. Theoretically, when $\gamma \rightarrow 0$, the optimization problem is ill-defined, while when $\gamma \rightarrow \infty$, the domain-invariant feature learning and discriminative graph self-learning are not performed, thus our CDGS can not learn robust features for cross-domain label propagation. As we can see from Fig. \ref{parameter_analysis} (b), determining the optimal value of $\gamma$ is infeasible and a reasonable one will make CDGS outperform the best competitor generally. Finally, we vary the value of $\beta$ from 0.001 to 10.0 to evaluate its influence. Theoretically, too small (large) values of $\beta$ make the label smoothness constraint (graph self-learning with the projected features) ineffective, which hinders us to construct a high-quality affinity matrix. A proper value of $\beta$ helps to capture the intrinsic similarity of samples, thereby improving the performance of cross-domain label propagation. From Fig. \ref{parameter_analysis} (c), we can discover that $\beta \in [0.01,5.0]$ is an optimal choice. Moreover, we display the convergence analysis in Fig. \ref{parameter_analysis} (d), where the maximum iteration is 15. We can observe that our CDGS can quickly converge within several iterations.

\subsection{Semi-supervised Domain Adaptation}
\subsubsection{Results on Office-Caltech10 dataset}
We follow the standard experimental setup of \cite{MMDT},  where 20 samples per class are randomly selected for amazon domain while 8 for the others as the sources. Besides, three labeled target samples per category are selected for training with the rest for testing. For fair comparison, we use the train/test splits released by \cite{MMDT}. The average accuracies for each task over 20 random splits are shown in Table \ref{office_surf_semi}. We also report the performance of OBTL \cite{OBTL}, which to our knowledge, is the best method on this dataset. We can observe that in terms of the average accuracy, CDGS obtains 2.2$\%$ improvement over OBTL. Besides, CDGS works the best for 9 out of all 12 tasks while OBTL just wins one task, which verifies the significant effectiveness of our semi-supervised extension. Carefully comparing the results of Table \ref{office_surf_semi} and Table \ref{Office_surf_split}, we find that when few labeled target samples are available, CDGS obtains 11.4$\%$ gain in the average classification performance, which highlights the value of our extension.  

\subsubsection{Results on MNIST-USPS dataset}
We follow the protocol of \cite{TFMKL-S}. Specifically, all source samples are utilized for training, and 2 labeled target samples per category are also selected for training with the remaining to be recognized. The average classification accuracies over 5 random splits are reported in Table \ref{mnist_semi}, where some results are copied from \cite{TFMKL-S}. We can observe that our CDGS is the best method for all tasks and achieves 83.5 $\%$ averaged accuracy, leading the second best method CDLS by 9.7$\%$, which confirms the superiority of our semi-supervised extension.

\section{Conclusion and Future Work \label{conclusion}}
In this paper,  a novel domain adaptation approach called CDGS is proposed, which infers target pseudo-labels by cross-domain label propagation. Different from existing cross-domain label propagation methods that separate domain-invariant learning, affinity matrix constructing and target labels inferring into three independent stages, 
our CDGS integrates these three parts into one unified optimization framework, such that they can assist each other to achieve more effective knowledge transfer. Furthermore, to construct a high-quality affinity matrix in CDGS, we propose a discriminative graph self-learning strategy, which can capture the inherent data manifold structure by adaptively calculating sample similarity in the projected space and exploring the discriminative information contained in well-labeled source data and pseudo-labeled target data. An iterative optimization algorithm is designed to solve the CDGS optimization problem. We further extend our CDGS to the SDA scenario in a direct but effective way and the corresponding optimization problem can be solved with the identical optimization algorithm. Extensive experimental results on six benchmark datasets have verified the significant superiority of our CDGS against the competitors in both UDA and SDA settings.

\ifCLASSOPTIONcaptionsoff
\newpage
\fi

\end{document}